\documentclass[10pt,twocolumn,letterpaper]{article}

\usepackage{longtable}
\usepackage[accsupp]{axessibility}

\usepackage{amssymb}
\usepackage[pagenumbers]{cvpr} 
 \usepackage{multirow}
 \usepackage{tabularx} 
\usepackage[dvipsnames,table]{xcolor}

\definecolor{cvprblue}{rgb}{0.21,0.49,0.74}
\usepackage[pagebackref,breaklinks,colorlinks,citecolor=cvprblue]{hyperref}

\newcommand{\hidden}[1]{}

\usepackage{tcolorbox}

\newcommand{\greentext}[1]{\textcolor[rgb]{0, 0.5, 0}{#1}}
\newcommand{\orangetext}[1]{\textcolor[RGB]{255, 69, 0}{#1}}
\newcommand{\bluetext}[1]{\textcolor[RGB]{0, 0, 255}{#1}}
\newcommand{\purpletext}[1]{\textcolor[RGB]{148, 0, 211}{#1}}

\newcommand{\newshortname}{ViP-LLaVA}
\newcommand{\newshortnamebase}{ViP-LLaVA-Base}
\newcommand{\newbenchmarkshortname}{ViP-Bench}

\def\paperID{14} 
\def\confName{CVPR}
\def\confYear{2024}

\title{\newshortname{}: \\Making Large Multimodal Models Understand Arbitrary Visual Prompts}

\author{
Mu Cai$^{1}$ \quad  Haotian Liu$^{1}$ \quad Dennis Park$^{2}$ \quad  
Siva Karthik Mustikovela$^{2}$ \quad  \\Gregory P. Meyer$^{2}$ \quad  Yuning Chai$^{2}$  \quad  Yong Jae Lee$^{1,2}$\\
$^{1}$University of Wisconsin--Madison~~~~~~$^2$Cruise LLC \\
\href{https://vip-llava.github.io}{https://vip-llava.github.io}
}

\begin{document}
\maketitle
\begin{abstract}
\vspace{-5pt}
While existing large vision-language multimodal models focus on whole image understanding, there is a prominent gap in achieving region-specific comprehension. Current approaches that use textual coordinates or spatial encodings often fail to provide a user-friendly interface for visual prompting. To address this challenge, we introduce a novel multimodal model capable of decoding arbitrary (free-form) visual prompts. This allows users to intuitively mark images and interact with the model using natural cues like a ``red bounding box'' or ``pointed arrow''. Our simple design directly overlays visual markers onto the RGB image, eliminating the need for complex region encodings, yet achieves state-of-the-art performance on region-understanding tasks like Visual7W, PointQA, and Visual Commonsense Reasoning benchmark. Furthermore, we present \newbenchmarkshortname{}, a comprehensive benchmark to assess the capability of models in understanding visual prompts across multiple dimensions, enabling future research in this domain. Code, data, and model are publicly available. 
\end{abstract}  
\vspace{-5pt}
\section{Introduction}
\label{sec:intro}

Large language models (LLMs) like ChatGPT~\cite{chatgpt}, GPT4~\cite{gpt4}, and Bard~\cite{Google_Bard} have recently gained significant attention for their strong reasoning and generalization capabilities, and their ability to chat in a human-like manner. In particular, models such as GPT-4V(ision)~\cite{GPT4V_System_Card}, which incorporate visual information, have demonstrated human-level perception and reasoning capabilities~\cite{yu2023mmvet}. This has spurred the development of similar open-source models that aim to replicate or even surpass the proprietary models' performance.

\begin{figure}[t]
\centering
\includegraphics[width=0.99\linewidth]{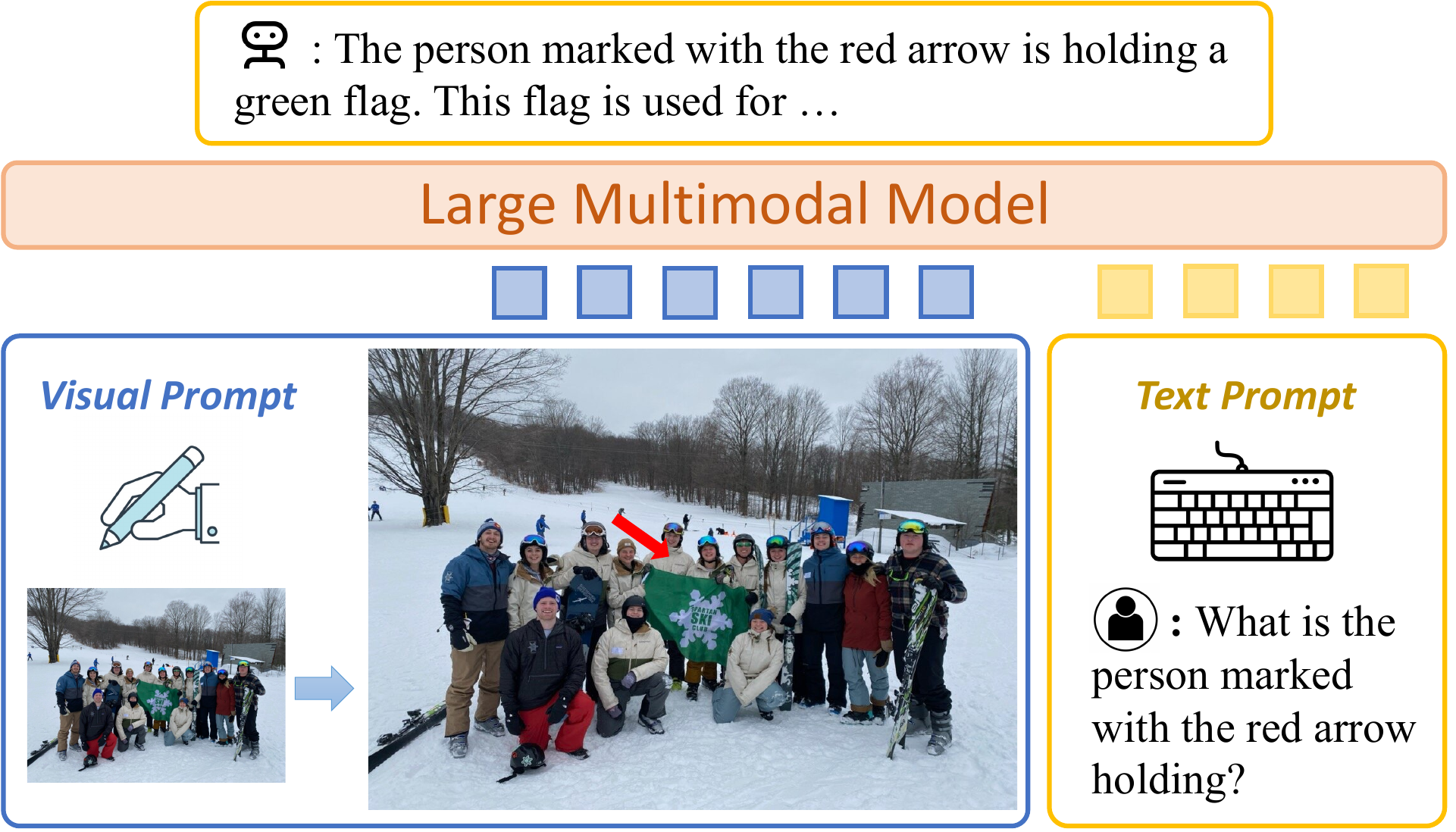}
\caption{
\textbf{Main Idea of \newshortname{}.}  We directly overlay diverse visual prompts (e.g., arrows, boxes, circles, scribbles) onto the original image, and then feed the corresponding visual features along with text embeddings into the large multimodal model for conversational assistance.  Here we show an example using a red arrow.
}
\vspace{-1em}
\label{fig:teaser}
\end{figure}

Despite their capabilities, current models, including seminal ones like LLaVA~\cite{liu2023llava, liu2023improvedllava} and MiniGPT-4~\cite{zhu2023minigpt}, focus predominantly on whole-image understanding; in other words, they lack the capability to process \emph{region-specific} information in complex scenes. This limitation becomes particularly apparent when attempting to describe specific objects within an image using only language prompts, which can be difficult when there is ambiguity (e.g., when there are multiple people in the image, and the question relates to a specific person), as shown in Figure~\ref{fig:teaser}.

To address this issue, recent work explores spatial references in multimodal models. Existing efforts have primarily focused on using textual representations of coordinates~\cite{zhao2023chatspot, ferraiolo2000scalable, chen2023shikra, chen2023minigptv2}, learned positional embeddings~\cite{zhang2023gpt4roi, peng2023kosmos, zhou2023regionblip}, or ROI features~\cite{zhang2023gpt4roi, rasheed2023glamm}. However, they often lack user-friendliness, as they are limited to fixed-format visual references like bounding boxes and the spatial coordinates of a mask contour. Most of these approaches, including those by Zhang et al.~\cite{zhang2023gpt4roi} and Chen et al.~\cite{chen2023shikra}, only employ bounding box inputs for visual referrals. While effective in structured scenarios, this method proves less versatile in natural, user-driven interactions where the visual prompts may not conform to clean geometric shapes.

In this paper, we propose a simple yet highly effective solution to this problem: a large multimodal model that can process \emph{arbitrary visual prompts}. This allows a user to intuitively mark up images and interact using natural cues such as a ``red bounding box'' or ``pointed arrow''. Our model recognizes these visual prompts, offering a user-friendly way to integrate visual references into the language dialogue.  Based on our own observation and prior work~\cite{shtedritski2023redcircle}, which shows that CLIP can understand visual markers, we \emph{directly inject the visual prompts into the original image space} without any additional region-specific model designs. Although our approach is deceptively simple, it yields an unexpected benefit: our model sets new state-of-the-art performances on tasks demanding precise region-specific perception and complex reasoning. It surpasses the capabilities of existing related models with specialized region encoding techniques, as evidenced by our superior performance on region reasoning tasks on Visual7W~\cite{zhu2016cvpr-v7w} and PointQA~\cite{mani2020point}.

To further support research in this area, we introduce \emph{\newbenchmarkshortname{}}, a benchmark for evaluating multimodal models' region understanding capabilities with arbitrary visual prompts. By collecting a diverse set of 303 images and questions, we provide a comprehensive assessment of visual understanding capabilities across six aspects at the region level: recognition,  OCR, knowledge, math, object relationship reasoning, and language generation. We believe that \newbenchmarkshortname{} will provide a solid foundation for future research into multimodal models with arbitrary visual prompts.

In summary, our main contributions are:
\begin{itemize}
    \item We introduce a novel multimodal model for intuitive interaction with images using natural language and arbitrary visual prompts, enhancing user accessibility and model flexibility.
    
    \item We develop a visual referal approach that overlays visual prompts directly onto images, simplifying the model's architecture without compromising performance.
    
    \item Our model, \newshortname{},  achieves state-of-the-art results on region understanding tasks on established benchmarks, surpassing specialized region encoding models.
    
    \item We introduce \newbenchmarkshortname{}, a benchmark for evaluating visual prompt interpretation, setting a foundational platform for future research.
\end{itemize}
\section{Related Work}
\label{sec:related_work}

\paragraph{Advancements in Large Multimodal Models.}
Large language models like ChatGPT~\cite{chatgpt}, GPT4~\cite{gpt4}, and LLaMA~\cite{touvron2023LLaMA} have shown impressive reasoning and generalization capabilities. The landscape of LLMs has been markedly transformed by the recent introduction of models that integrate visual information, such as GPT-4V(ision)~\cite{GPT4V_System_Card}.  Building upon open-source LLMs~\cite{touvron2023LLaMA,Vicuna}, a vast number of multimodal vision-language models have made significant strides, spearheaded by LLaVA~\cite{liu2023llava, liu2023improvedllava} and MiniGPT-4~\cite{zhu2023minigpt}, which combine LLaMA's~\cite{touvron2023LLaMA} language prowess with a CLIP~\cite{radford2021learning} based image encoder.  
While these models excel at whole-image understanding, a key challenge has been region-specific comprehension within complex visual scenes. This has led to the exploration of spatial referrals in multimodal contexts. Existing models utilize textual coordinate representations~\cite{zhao2023chatspot, ferraiolo2000scalable, chen2023shikra, chen2023minigptv2}, learned positional embeddings~\cite{zhang2023gpt4roi, peng2023kosmos, zhou2023regionblip}, or Region of Interest (ROI) features~\cite{zhang2023gpt4roi} to anchor language to specific image regions. However, they often employ rigid visual referral formats that are not as intuitive for users.

\vspace{-10pt}
\paragraph{Visual Prompting as a User-Friendly Solution.}
Our focus is on making the interaction with multimodal models more natural and intuitive. Traditional models have employed regular shapes for visual prompting, but our research is motivated by the need for a system that can interpret a wider range of visual prompts. For example, in visual perception, interactive segmentation methods have been proposed that can take in points or scribbles~\cite{zou2023segment, kirillov2023segment}.  Drawing inspiration from recent findings that show GPT-4V's ability to understand a variety of markers~\cite{yang2023dawn}, we advocate for a model that can handle arbitrary visual cues, such as scribbles and arrows.  In our model,  \newshortname{}, we overlay these visual prompts directly onto the image canvas. This is accomplished by fine-tuning on a dataset specifically designed for arbitrary visual prompt instructions.

\vspace{-10pt}
\paragraph{Evaluating LMM's Region Understanding Capabilities.}

{Existing works~\cite{chen2023shikra,zhang2023gpt4roi, peng2023kosmos, you2023ferret} evaluates the model's region understanding capabilities on regional multichoice~\cite{zellers2019vcr, zhu2016cvpr-v7w, mani2020point} or  captioning~\cite{yu2016modeling-refcoco, krishna2017visual} tasks with metrics such as accuracy, recall, and CIDer~\cite{vedantam2015cider}. However, these metrics fall short when it comes to evaluating visual dialogue for large multimodal models in an open-world setting.
To evaluate LMM's capability in engaging in visual conversations for \emph{image-level} understanding, two families of evaluation are proposed: multiple-choice~\cite{liu2023mmbench} or using GPT4 as a judge for free-form answers~\cite{yu2023mmvet, liu2023llava}.
However, a gap still exists in the evaluation of LMM's capabilities for comprehending arbitrary visual prompts. To address this, we introduce \newbenchmarkshortname{}, a comprehensive benchmark tailored to evaluate how well the LMMs can interpret various visual prompts across multiple dimensions, including recognition, OCR, knowledge, math, relationship reasoning, and language generation.
\section{Approach}

Our research hinges on the premise that a large multimodal model should not only perceive the visual content of an image but also interpret arbitrary visual markers as part of the user interaction. In this section, we describe our approach that achieves this goal, highlighting the pivotal role of CLIP in understanding visual markers and the construction of a new instruction tuning dataset tailored to train \newshortname{} to understand arbitrary visual prompts.

\subsection{Visual Prompt Embedding via CLIP}
\label{sec:clip-marker}

In contrast to prior work on region understanding~\cite{peng2023kosmos, zhang2023gpt4roi} which constructs a new module to process visual prompts, we leverage CLIP's~\cite{radford2021learning} existing capabilities to encode both the image and superimposed visual markers. Specifically, CLIP's proficiency in aligning visual and textual data makes it an ideal candidate for this task, as recent studies~\cite{shtedritski2023redcircle} suggest that it inherently pays attention to marked regions including circles, rectangles, \etc. As shown in our experiments, we further demonstrate that CLIP can focus the model's attention on a wider variety of visual prompts such as arrows and arbitrary scribbles. To utilize this functionality, we composite the visual prompts $\mathbf{P}_{\mathrm{v}}$ onto the original image $\mathbf{X}_{\mathrm{v}}$ through alpha blending, creating a merged representation that highlights the areas of interest:
\begin{equation}
\hat{\mathbf{X}}_{\mathrm{v}} = \alpha \cdot \mathbf{P}_{\mathrm{v}} + (1 - \alpha) \cdot \mathbf{X}_{\mathrm{v}},
\end{equation}
where $\alpha \in [0,1]$ denotes the transparency level of the visual prompt, $\mathbf{X}_{\mathrm{v}}$ is the image, and $\mathbf{P}_{\mathrm{v}}$ is the image with the visual prompt.  Note that we only perform alpha blending for pixels underlying the visual prompt. The composite image $\hat{\mathbf{X}}_{\mathrm{v}}$ is then fed into the multimodal model.

\begin{figure}[t]
\centering
\includegraphics[width=0.99\linewidth]{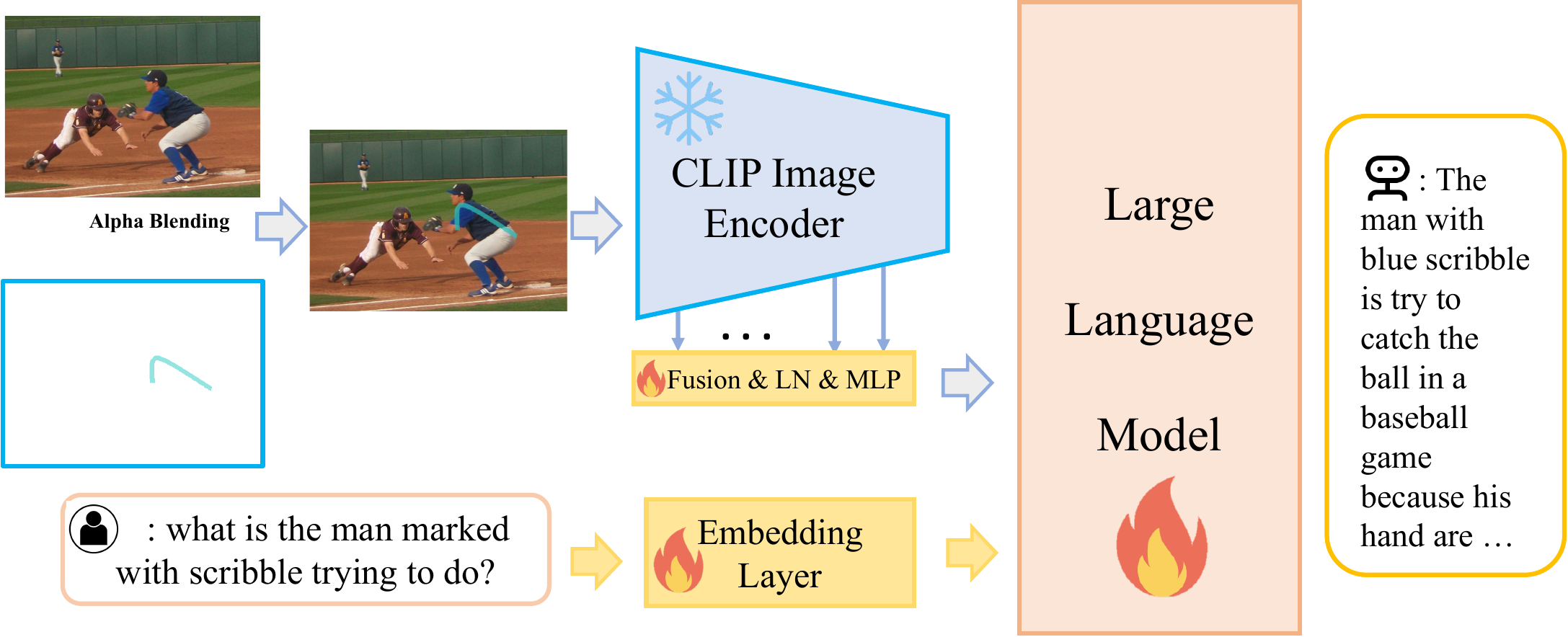}
\caption{\textbf{Model Architecture.} After alpha blending the visual prompts onto the original image, we feed the resulting image into the visual encoder to obtain multi-level visual features. Those features are concatenated and fed into the LayerNorm and MLP layers to form the visual tokens. Then visual tokens and text instruction tokens are fed into the large language model to produce the language response in an auto-regressive manner. The frozen and trainable modules during instruction tuning are annotated. 
}
\vspace{-0.5em}
\label{fig:approach}
\end{figure}

To effectively recognize the visual prompts, we balance low-level and high-level visual features in \newshortname{}.

To address the tendency of CLIP's deeper features to overlook low-level details~\cite{zhou2022maskclip}, we selectively extract features from multiple CLIP layers. Specifically, we use one early layer (6-th) to encode detailed geometric shapes and four deeper layers (15, 18, 21, 24-th) to capture broader semantic information.
These multi-level features are then concatenated, normalized using LayerNorm~\cite{ba2016layer} for training stability, and finally passed through an MLP layer. This process ensures \newshortname{} effectively integrates diverse visual cues, a strategy validated through our ablation studies detailed in Sec.~\ref{sec:ablation}.

Our design's simplicity of directly overlaying visual prompts offers several advantages. It reduces model complexity by bypassing additional processing modules and aligns closely with natural human interactions, as users often employ diverse and spontaneous visual markers. This flexibility allows \newshortname{} to interpret a wide range of user-generated visual cues, enhancing its applicability in real-world scenarios.

To train \newshortname{}, we perform autoregressive language modeling; \ie, we maximize the likelihood of generating the tokens of the ground-truth answer $\mathbf{X}_{\mathrm{a}}$:
\begin{equation}
P(\mathbf{X}_{\mathrm{a}} \mid \hat{\mathbf{X}}_{\mathrm{v}}, \mathbf{X}_{\text {instruct}})=\prod_{i=1}^L P_{\boldsymbol{\theta}}(x_i \mid \hat{\mathbf{X}}_{\mathrm{v}}, \mathbf{X}_{\text {instruct}}, \mathbf{X}_{\mathrm{a},<i})
\end{equation}
where $\boldsymbol{\theta}$ represents the trainable parameters, $\mathbf{X}_{\text {instruct}}$ is the text instruction, $L$ is the sequence length of the answer $\mathbf{X}_{\mathrm{a}}$, and $\mathbf{X}_{\mathrm{a},<i}$ denotes all the answer tokens before the current prediction token $x_i$, where $i$ denotes the steps during text token generation. Here we omit system messages from the equation for clarity, even though they are part of the conditioning. Figure~\ref{fig:approach} shows our model architecture.

This training objective enables the model to generate contextually accurate responses by comprehending the visual content, language instruction, and the overlaid prompts. It fosters the model's ability to interpret visual markers in unison with the image, thereby enhancing its proficiency in addressing complex, region-specific language inquiries. This capability is crucial for tasks requiring nuanced understanding of both the visual elements and user intentions conveyed through arbitrary visual prompts.

\subsection{Visual Prompting Design}\label{sec:visualpromptingdesign}

To train the model to recognize and interpret arbitrary visual prompts, we develop a new visual prompt instruction tuning dataset, as there are no prior datasets with arbitrary visual prompts and instruction-output text pairs that we can use. 

Our dataset comprises a diverse collection of 520k image-text pairs marked with visual prompts, sourced from publicly available datasets, including (1) single region reasoning data: 80k referring comprehension and generation data from RefCOCOg~\cite{yu2016modeling-refcoco}, and 37k object counting data from PointQA-LookTwice~\cite{mani2020point}, (2) two-region reasoning data: 80k triplet relationship data from Visual Genome~\cite{krishna2017visual}, (3) multi-region reasoning data: 30k grounded image captioning data from Flicker 30k Entities~\cite{plummer2015flickr30k}, 213K data from Visual Commonsense Reasoning dataset~\cite{zellers2019vcr},  and 82k data from Visual7W~\cite{zhu2016cvpr-v7w}.  Note that all those data are collected from the training split of the aforementioned datasets.

We automatically annotate each image with various visual prompts. For the data that only comes with bounding box annotations, we sample the visual prompts from three possible categories: rectangle, ellipse, and arrow. Here we make sure that the head of the arrow lies within $[(- \frac{W}{2}, - \frac{H}{2}), (\frac{W}{2}, \frac{H}{2})]$ space, where $W, H$ are the width and height of the image, respectively. For ellipse, the lengths along the semi-major and semi-minor axes are inherited from the bounding box size, where we enlarge the ellipse with a ratio between $[1, 1.5]$.  On the other hand, for regions that come with ground truth pixel-level mask annotations, we annotate each region with visual prompts sampled from the following 8 possibilities: rectangle, ellipse, point, triangle, mask, mask contour, arrow, and scribble created using Bézier curves; see Figure~\ref{fig:vis_visual_prompts}.  We make sure that the head of the arrow, entire point, triangle, and scribble lies within the provided mask.  These annotations simulate natural human interactions with images, where users often use spontaneous markers to highlight areas of interest.

For scribbles, we simulate human-like drawings using Bézier curves~\cite{farin2014curves}. This process begins by randomly selecting three points within the object mask, which serve as the anchors for the quadratic Bézier curve. The generated Bézier curve is then composited onto the image using the previously mentioned alpha blending technique to produce a merged image with the scribble serving as a visual prompt.

\begin{figure}[t]
\centering
\includegraphics[width=0.99\linewidth]{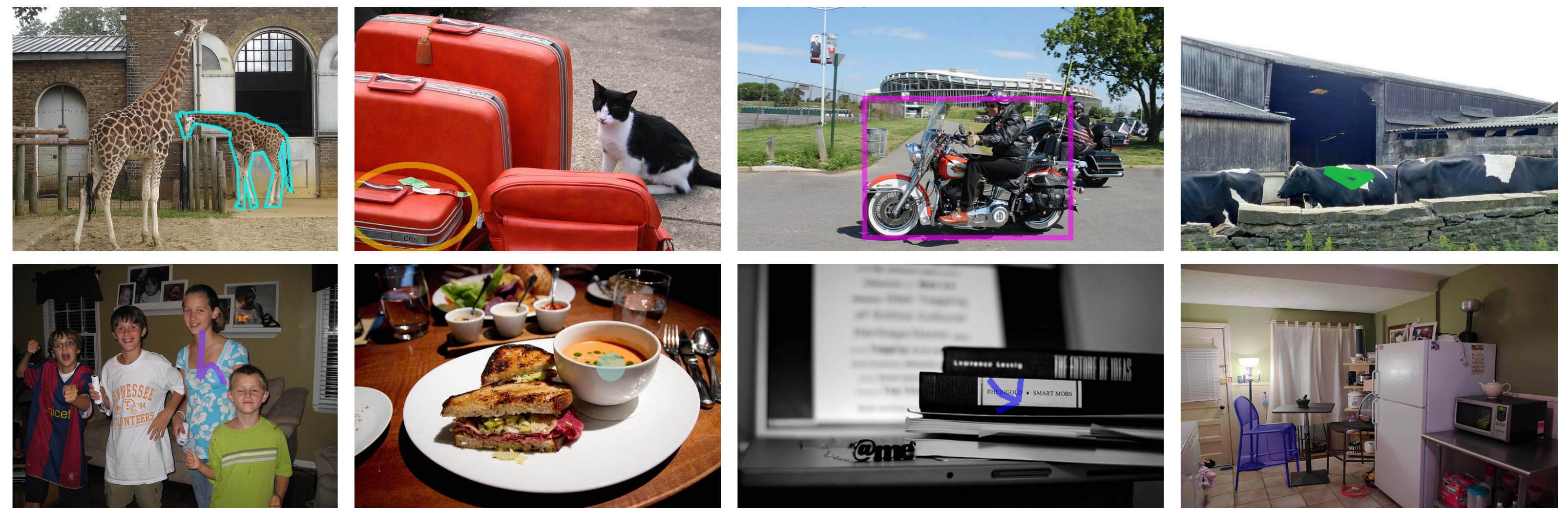}
\caption{\textbf{Visualization of Visual Prompt Types.} From top-left to bottom-right: mask contour, ellipse, bounding box, triangle, scribble, point, arrow, and mask. Note that the prompts not only have diverse shapes, but they also have diverse colors, transparency values, widths, scales, and directions.}
\vspace{-1.5em}
\label{fig:vis_visual_prompts}
\end{figure}

Humans naturally use various markers to highlight objects within their environment. For instance, in educational settings, teachers often use arrows or underlining to draw students' attention to specific parts of an image or text. Similarly, in everyday communication, people might circle items in a photograph to point out something of interest or use scribbles to obscure sensitive information before sharing.  Through our design, we create a visual instruction following dataset that mirrors the way humans visually interact with objects, thus fostering a more intuitive and natural interaction with the model.

\subsection{Optional Region-level Instruction Tuning Data}
\label{sec:gpt4v-data}

Our training data comes from two sources: (i) region-level visual prompting data described in Section~\ref{sec:visualpromptingdesign}, and (ii) image-level data devoid of visual prompts, sourced from LLaVA-1.5~\cite{liu2023improvedllava}. This strategy enables \newshortname{} to engage in human-like conversations, primarily due to the image-level LLaVA instruction data from Liu \etal~\cite{liu2023llava}.  Optionally, to  further enhance \newshortname{}'s capability in multimodal conversations at the region-level, we design region-specific instruction data with the help of GPT-4V. 

Prior approaches like Shikra~\cite{chen2023shikra} attempted to generate region-level instruction data using text-only models like GPT4. However, this method is inherently limiting, particularly in object-level tasks where the model, lacking visual context, cannot accurately reference multiple objects of the same class within a single scene.  To overcome this, we develop an instruction data curation method using GPT-4V. Unlike text-only models, GPT-4V can interpret visual prompts displayed in images~\cite{yang2023dawn}. Our method involves feeding two images into GPT-4V: the original image and a modified version with annotated visual prompts. Alongside these images, we provide the model with the ground-truth (text) annotation in the original dataset and system messages. This process is used to curate \texttt{<visual prompt, text prompt, text output>} triplets for the images in our dataset described in Section~\ref{sec:visualpromptingdesign}.

We introduce specific textual representations such as \texttt{<within red mask>} and \texttt{(<within red box>, <within blue box>)} to guide GPT-4V in recognizing the visual prompts in both single-region and multi-region settings. During training, we replace these phrases with the set of eight possible visual prompts described in Section~\ref{sec:visualpromptingdesign}, significantly enhancing the dataset's versatility. In total, we curate 13k high-quality region-level instruction data points, comprised of 7k single-region and 6k multi-region instances. In the supplementary, we provide specific details of the system messages, input text prompts, and generated text outputs.

Although \newshortname{} works well even without this enriched data for standard visual reasoning benchmarks, we find that it helps to further improve the model's ability to have human-like conversations in open-world settings.

\section{\newbenchmarkshortname{} for Evaluation}

In order to rigorously evaluate the capabilities of multimodal models in interpreting and responding to visual reasoning queries, we introduce \newbenchmarkshortname{}, a benchmarking suite for evaluating multimodal region-understanding capabilities under various visual prompts. \newbenchmarkshortname{} consists of 303 unique image-question pairs, where images are collected from MM-Vet~\cite{yu2023mmvet}, MMBench~\cite{liu2023mmbench}, and Visual Genome~\cite{krishna2017visual}. Each pair consists of an image coupled with a diverse visual reasoning question designed to test a model's understanding and interpretation capabilities. We reuse the questions in MM-Vet~\cite{yu2023mmvet} and MMBench~\cite{liu2023mmbench} (but make minor adjustments so that they take into account the region-specific visual prompts), while in Visual Genome, we design the questions and answers by ourselves. We use bounding boxes and masks produced by the Segment Anything Model~(SAM)~\cite{kirillov2023segment} to annotate the location of the objects.

Key to the design of \newbenchmarkshortname{} is its comprehensive coverage of six crucial aspects of visual understanding at the region level: recognition,  OCR (Optical Character Recognition), knowledge, math, object relationship reasoning, and language generation. This range ensures a holistic assessment of a model's performance in various facets of region-level visual reasoning.

\newbenchmarkshortname{} employs a similar grading mechanism as MM-Vet~\cite{yu2023mmvet}. We employ the GPT-4 text model, a state-of-the-art language model, to evaluate the responses of multimodal models. Specifically, we feed the response from the multimodal model, the human annotated answer, and several in-context scoring examples to GPT-4. The responses are scored by GPT-4 on a scale from 0 to 10, offering a quantitative measure of the multimodal model's proficiency in understanding and interpreting visual data. This grading system provides a standardized framework for comparing the performance of different models. 

\newbenchmarkshortname{} is meticulously annotated by humans. This process involved seven rounds of validation to ensure the accuracy and relevance of the object boxes/masks, questions, and answers. Such rigorous annotation guarantees the reliability of the benchmark as a tool for model evaluation.  An illustrative example in Table~\ref{fig:regionbench} showcases a scenario where a leading model like GPT-4V misinterprets object localization under \newbenchmarkshortname{}, highlighting the challenges in current multimodal understanding. We present additional visualizations and statistics of \newbenchmarkshortname{} in the supp.

Through \newbenchmarkshortname{}, we provide a valuable tool for the research community, aiding in the development and refinement of multimodal models. By offering a comprehensive and challenging testbed, we believe \newbenchmarkshortname{} can set the stage for future advancements in the field of visual reasoning and multimodal interaction.

\begin{table}[t]
\centering
\footnotesize
\begin{tabular}{@{}lcc@{}}
\toprule
Method         & Generalist? & Accuracy (\%) \\ \midrule
LSTM-Att~\cite{zhu2016cvpr-v7w}       & $\times$  & 56.10          \\
CMNs~\cite{hu2017modeling}           & $\times$  & 72.53         \\
12in1~\cite{lu202012}          & $\times$  & 83.35         \\
GPT4ROI-7B ~\cite{zhang2023gpt4roi}     & $\times$  & 81.83         \\
GPT4ROI-13B~\cite{zhang2023gpt4roi}  & $\times$  & 84.82         \\
Shikra-13B~\cite{chen2023shikra} & \checkmark  & 85.33         \\ \hline
\newshortnamebase{}-7B   & \checkmark  & {86.04}  \\  
\newshortnamebase{}-13B   & \checkmark  & {87.54}  \\  
\newshortname{}-7B   & \checkmark  & {86.60}  \\   
\newshortname{}-13B    & \checkmark  & \textbf{87.91}  \\   \bottomrule
\end{tabular}
\caption{Comparison of methods in terms of generality and accuracy on Visual7W~\cite{zhu2016cvpr-v7w} test set.}
\vspace{-1em}
\label{tab:v7w}
\end{table}

\section{Experiments}

In this section, we compare \newshortname{} to state-of-the-art multimodal models, including those that explicitly design region-specific modules, perform in-depth analysis to assess \newshortname{}'s capabilities, and perform ablation studies.   

\begin{table}[t]
\centering
\footnotesize
\label{tab:method_comparison}
\begin{tabular}{lcc}
\toprule
Method         & Generalist? & Accuracy (\%) \\ \midrule
Point and ask~\cite{mani2020point}  & $\times$    & 60.20         \\
LLaVA-1.5-7B~\cite{liu2023improvedllava}      & \checkmark  & ~~56.19$^\dagger$         \\
LLaVA-1.5-13B~\cite{liu2023improvedllava}      & \checkmark  & ~~57.93$^\dagger$         \\
Shikra-13B~\cite{chen2023shikra}     & \checkmark  & 70.30          \\ \hline
\newshortnamebase{}-7B      & \checkmark  & {70.86 }     \\ 
\newshortnamebase{}-13B  & \checkmark  & \textbf{72.15 }  \\   
\newshortname{}-7B      & \checkmark  &      {71.31}  \\  
\newshortname{}-13B      & \checkmark  & {71.77 }     \\  \bottomrule
\end{tabular}
\caption{Comparison of methods in terms of generality and accuracy on PointQA-LookTwice~\cite{mani2020point} test set. $^\dagger$zero-shot eval.}
\vspace{-0.5em}
\label{tab:pointqa_looktwice}
\end{table}

\subsection{Training Setup}

\paragraph{Model.} For the visual model, we choose CLIP-336px~\cite{radford2021learning} to preserve more information from the raw pixel space. We use Vicuna v1.5~\cite{Vicuna} as the language encoder. For the multimodal connector, a 2-layer MLP is utilized. Ablations on more LLM backbones are shown in Supp.~\ref{sec:impact llm backbone}.

\vspace{-10pt}
\paragraph{Training and data.} During the initial stage of training, we employ 558k BLIP~\cite{instructblip, liu2023llava} captioned image-text pairs to pretrain the multimodal connector. The second stage utilizes LLaVA v1.5~\cite{liu2023improvedllava} instruction data alongside our region-level visual prompting dataset from Section~\ref{sec:visualpromptingdesign}. Both stages train the model for 1 epoch, with an overall training time of around 20/40 hours for the 7B/13B model using 8 NVIDIA A100 GPUs. Finally,  we mix the 13k GPT-4V instruction data with 13k  sampled data from stage 2 to get 26k stage 3 training data, and then fine-tune our stage-2 model (referred to as \newshortnamebase{}) for one epoch to get our model \newshortname{}, which requires approximately 0.5 hours for the 7B model and 1 hour for the 13B  model on 8 NVIDIA A100 GPUs.

\vspace{-10pt}
\paragraph{Visual prompts.} \newshortname{} uses 8 visual prompts: rectangles, ellipses, points, scribbles, triangles, masks, mask contours, and arrows. Their attributes, such as color, thickness, and alpha value for alpha blending (in [0.5, 1]) are randomized. The arrow's direction and length are randomized, with the endpoint remaining within the mask. For referencing specific regions, we replace the \texttt{<region>} text with the color and shape description, such as \texttt{red scribble}. The visual prompt type and associated attributes for each region are randomly assigned during training.

\subsection{Evaluation on Region Reasoning Benchmarks}

We first quantitatively evaluate \newshortname{} on three region reasoning benchmarks.

\vspace{-10pt}
\paragraph{Visual7W.} The Visual7W dataset~\cite{zhu2016cvpr-v7w} tests models' spatial perception by requiring them to match text descriptions with the correct bounding boxes from a set of choices. We differentiate between `generalist' models, which are not specifically trained on the target dataset, and `specialist' models, which are.  For a fair comparison, we use image overlays as visual prompts for the LLaVA model and textual coordinates for Shikra's text prompts. The results in Table~\ref{tab:v7w} shows \newshortname{}-7B outperforming recent state-of-the-art methods, including GPT4RoI~\cite{zhang2023gpt4roi} and Shikra~\cite{chen2023shikra}, despite having fewer parameters, and \newshortname{}-13B producing even higher gains. \newshortname{} overlays bounding boxes directly onto the image, creating an immediate link between the image and spatial locations. This contrasts with other methods that rely on external embeddings from either textual or newly learned embedding spaces to reference specific regions, proving less effective in this context.

\begin{table}[t]
\centering
\footnotesize
\vspace{-3mm}
\begin{tabular}{lccc}
\toprule
Model & $Q \rightarrow A$ (\%) & $QA \rightarrow R$ (\%) & $Q \rightarrow AR$ (\%) \\
\midrule
ViLBERT~\cite{lu2019vilbert} & 72.4 & 74.5 & 54.0 \\
Unicoder-VL~\cite{li2020unicoder}  & 72.6 & 74.5 & 54.5 \\
VLBERT-L~\cite{su2019vl}  & 75.5 & 77.9 & 58.9 \\
ERNIE-ViL-L~\cite{yu2021ernie}  & 78.52 & 83.37 & 65.81 \\
VILLA-L~\cite{gan2020large}  & 78.45 & 82.57 & 65.18 \\
GPT4RoI-7B~\cite{zhang2023gpt4roi} & {87.4} & {89.6} & {{78.6}} \\
\newshortnamebase{}-7B &{\textbf{87.66}} &\textbf{89.80} & \textbf{78.93}\\ 
\bottomrule
\end{tabular}
\caption{Validation Accuracy on VCR~\cite{zellers2019vcr} dataset.}

\vspace{-1em}
\label{tab:vcr}
\end{table}

\paragraph{PointQA-LookTwice.} PointQA~\cite{mani2020point} presents a dataset where queries are based on either a specific point or a bounding box within an image. We evaluate \newshortname{} under the broad-question scenario using the bounding box type, typified by the prompt \texttt{How many of these are there?} This requires the model to first correctly identify the object within the given region and subsequently enumerate instances of the same category across the image---essentially a test of object recognition followed by class-specific counting. In line with our methodology for Visual7W, we use the image overlaid with the bounding box for LLaVA, while for Shikra, we incorporate the bounding box coordinates into the text prompt. Table~\ref{tab:pointqa_looktwice} shows \newshortname{}'s superior performance on this intricate task, surpassing other multimodal contenders. Our method of overlaying visual prompts ensures the object remains unobscured, effectively combining the original image pixels with visual cues to enhance object recognition and counting accuracy.

\vspace{-10pt}
\paragraph{Visual Commonsense Reasoning.}
The Visual Commonsense Reasoning (VCR) dataset~\cite{zellers2019vcr} is a challenging benchmark designed to evaluate a model's capabilities in high-level cognition and commonsense reasoning in the context of visual information. The dataset presents multiple-choice questions that require an understanding of the scene depicted in an image. Each question (Q) is paired with four potential answers (A), where the model must not only select the correct answer but also provide a rationale (R) that justifies its choice, demonstrating the model's ability to comprehend and rationalize visual elements within a given context.

We finetune \newshortnamebase{}-7B on VCR, similar to the protocol in GPT4RoI~\cite{zhang2023gpt4roi}. As shown in Table~\ref{tab:vcr}, our approach exhibits state-of-the-art performance on the validation set, illustrating its proficiency in visual commonsense reasoning. This success highlights our approach's dual strengths: adeptness in perception tasks and effectiveness in multi-region reasoning. By integrating visual prompts directly into the image, our model more effectively associates spatial locations with semantic understanding, facilitating a better interaction between spatial and semantic reasoning.

\begin{figure}[t]
\centering
\includegraphics[width=0.35\textwidth]{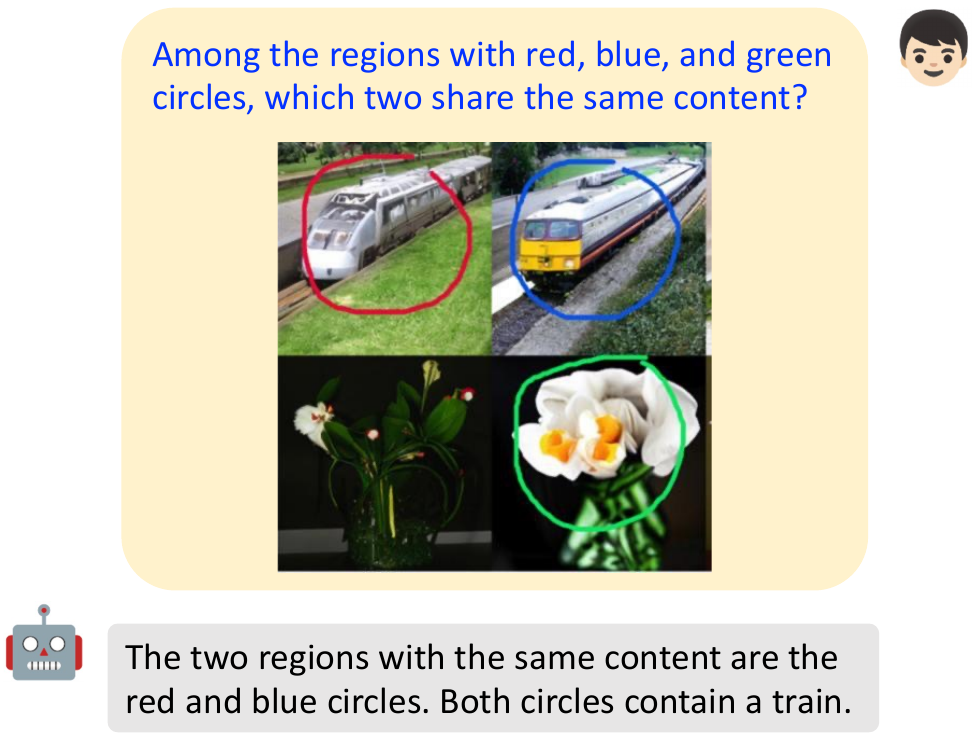}
\caption{\newshortname{} model is able to infer correspondences between multiple objects in the image.}
\label{fig:train}
\vspace{-0.5em}
\end{figure}

\begin{figure}[t]
\vspace{-1pt}
\centering
\includegraphics[width=0.5\textwidth]{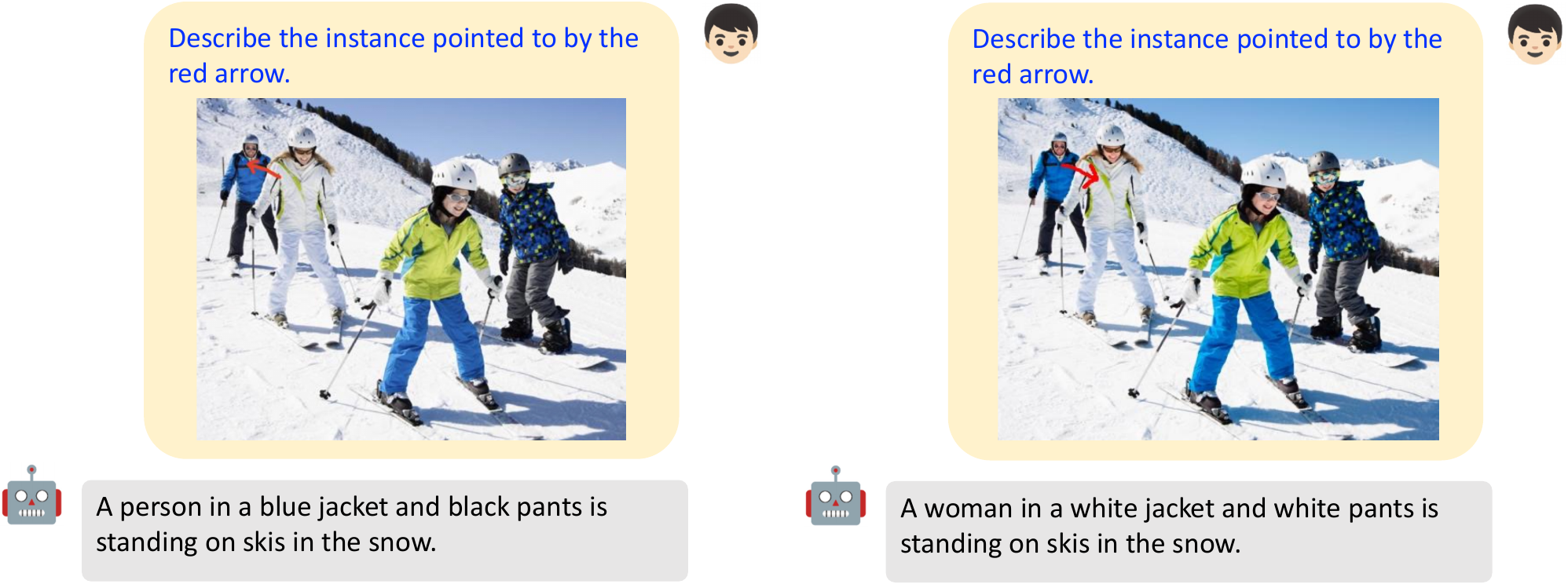}
\caption{\newshortname{} is able to understand the direction of arrows.}
\vspace{-0.5em}
\label{fig:arrow-compare}
\end{figure}

\subsection{In-depth Analysis}

\paragraph{Region reasoning with arbitrary human drawings at test time.} \newshortname{}, when presented with arbitrarily drawn enclosed regions or arrows by a user, can accurately describe, shown in Figure~\ref{fig:train} and~\ref{fig:dawn}.

\paragraph{Multi-region understanding capabilities.} \newshortname{} demonstrates robust multi-region understanding, able to dissect complex visual scenes and infer relationships between various elements. As shown in Figure~\ref{fig:train}, \newshortname{} is able to infer correspondences between multiple objects in the image, and make the correct reasoning that the red and blue circles both include the train.

\vspace{-10pt}
\paragraph{Arrow direction understanding.} \newshortname{} is able to understand arrows. Here we conduct an ablation study of the arrow direction. Given two arrows that have the same body yet different heads, as shown in Figure~\ref{fig:arrow-compare}, \newshortname{}  is able to understand the direction of the arrows, making correction descriptions about the respective regions.

\vspace{-10pt}
\paragraph{Generalization to other attributes.} 
\newshortname{} also generalizes to untrained attributes, like varying visual prompt thickness or location, showcasing its adaptability beyond what was seen during training. See the supplementary material for examples of different thicknesses. 

Figure~\ref{fig:dawn} further shows that \newshortname{} is able to conduct OCR first, and then make correspondences between different regions to make a correct prediction about the content of each part.

\begin{table}[t]
\centering
\footnotesize
\vspace{-3mm}
\begin{tabularx}{\columnwidth}{Xccc}
\toprule
Model & Visual7W & PointQA & VCR ($Q \rightarrow A$) (\%) \\
\midrule
Single Layer & 82.44 & 70.23 & 84.64 \\
Multi-Layer & \textbf{86.04} & \textbf{70.86} & \textbf{87.66} \\
\bottomrule
\end{tabularx}
\caption{Single layer vs.~multi-layer CLIP image encodings.}
\label{tab:ablation_feature}
\end{table}

\begin{figure}[t]
\centering
\includegraphics[width=0.7\linewidth]{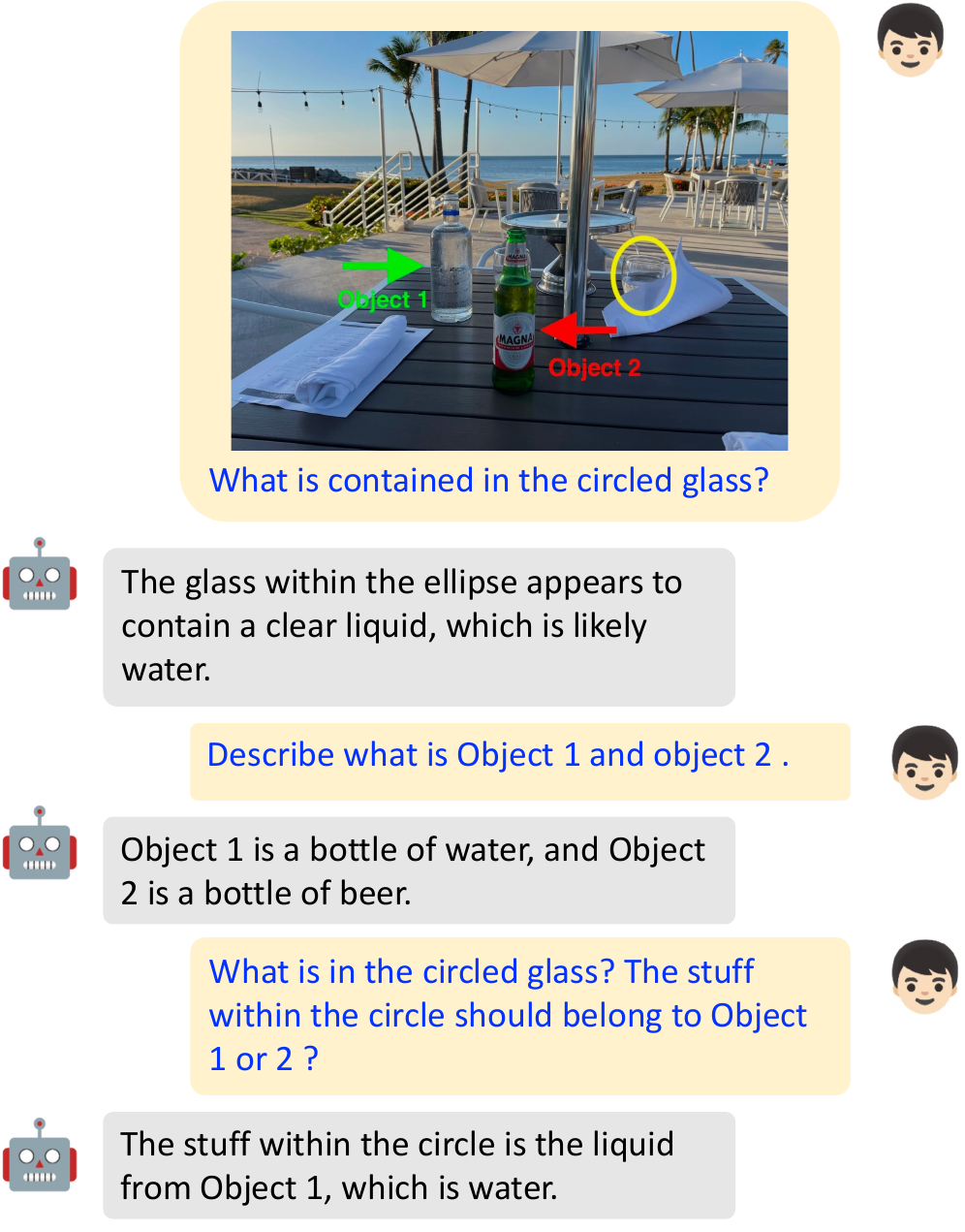}
\vspace{-1mm}
\caption{\newshortname{} is able to understand the text markers in the image, and reason about them.}
\vspace{-1em}
\label{fig:dawn}
\end{figure}

\begin{table*}[t]
\centering
\footnotesize
\begin{tabular}{ll|p{4mm}p{4mm}p{4mm}p{4mm}p{4mm}p{4mm}>{\columncolor{gray!20}}p{5mm}|p{4mm}p{4mm}p{4mm}p{4mm}p{4mm}p{4mm}>{\columncolor{gray!20}}p{5mm}}
\toprule
\multirow{2}{*}{Model} & \multirow{2}{*}{Format} & \multicolumn{7}{c|}{Synthesized visual prompts (tight bounding box)} & \multicolumn{7}{c}{Visual prompts from human (\eg arrow, circle)} \\
 &  & Rec & OCR & Know  & Math & Rel & Lang & All & Rec & OCR & Know  & Math & Rel & Lang & All \\ \midrule

GPT-4V-turbo-detail:high~\cite{gpt4} & VP &58.1 &   69.8  &  59.5  &  71.0 &   61.4  &  51.9  &  60.7 & 56.9  &    69.7  &    63.7  &    80.6 &     61.1 &     45.6   &    59.9   \\

GPT-4V-turbo-detail:low~\cite{gpt4} & VP & 53.2 &   50.3 &   55.6  &  67.7 &   57.5 &   57.5  &   52.8 & 51.7  &  50.3  &  59.3  &  60.3  &  55.0 &   43.8    &   51.4  \\

\midrule

InstructBLIP-7B~\cite{instructblip} & VP & 36.9 & 16.3 & 34.2 & 22.3 & 26.8 & 7.5 & 31.7 & 38.9 & 17 & 35.4 & 9.7 & 29.3 & 17.5 & 33.3 \\
{Shikra 7B }~\cite{chen2023shikra} & Coor &40.2	 & 10.0	 & 28.0 & 3.5	 & 18.9	 & 20.6 & 33.7   & -- & -- & -- & -- & -- & --  & --  \\
{GPT4ROI 7B }~\cite{zhang2023gpt4roi} & ROI &35.6 & 	16.7	 & 29.7	 & 9.7 & 	32.5 & 	13.8 & 35.1   & -- & -- & -- & -- & -- & --  & --  \\
Kosmos-2~\cite{peng2023kosmos} & Dis & 29.5 & 	14.2 & 	18.5	 & 9.7 & 	7.5	 & 21.9 & 26.9 & -- & -- & -- & -- & -- & --  & --  \\
LLaVA-1.5-7B~\cite{liu2023llava}  & Coor & 52.7	& 20.7	& 
 44.7 & 	14.5 & 	44.6 & 	30.6 & 	44.8 &  -- & -- & -- & -- & -- & -- & -- \\
LLaVA-1.5-7B~\cite{liu2023llava}  & VP & 50.8 & 12.4 & 49.2 & 6.5 & 51.8 & 23.8 & 41.6 & 49.1 & 13 & 42.9 & 9.7 & 50 & 27.5 & 40.2 \\
Qwen-VL-Chat~\cite{Qwen-VL} & Coor & 52.6 & 22.0 & 40.0 & 12.9 & 47.1 & 26.9 & 45.3 & -- & -- & -- & -- & -- & -- & -- \\
Qwen-VL-Chat~\cite{Qwen-VL} & VP & 43.0 & 30.4 & 40.2 & 9.7 & 25.7 & 28.7 & 39.2 & 48.7 & 22.1 & 41.2 & 6.5 & 48.2 & 25 & 41.7 \\
\newshortnamebase{}-7B & VP & 54.8 & 18.8 & 52.9 & 9.7 & 53.9 & 42.5 & {45.5} & 55.3 & 17.6 & 45.9 & 8.1 & 44.6 & 33.1 & {46.8} \\

\newshortname{}-7B  & VP & 56.7 & 	19.4 & 	49.7	 & 10.0	 & 50.4	 & 33.8 & \textbf{48.4} & 56.7	 & 21.2 & 	47.1 & 	12.3	 & 50.4	 & 36.2	 & \textbf{48.3} \\

\midrule
InstructBLIP-13B~\cite{instructblip} & VP & 42.5 & 12.2 & 37.5 & 3.2 & 33.2 & 12.5 & 35.8 & 41.7 & 13.6 & 35.9 & 3.2 & 27.9 & 18.8 & 35.2 \\
LLaVA-1.5-13B~\cite{liu2023llava} & Coor & 53.2 & 	26.1 & 	45.9 & 	9.7	 & 52.5 & 	31.9 & 	47.1 & -- & -- & -- & -- & -- & -- & -- \\
LLaVA-1.5-13B~\cite{liu2023llava} & VP & 48.1 & 21.8 & 40.0 & 6.1 & 45 & 28.1 & 41.8 & 48.8 & 21.3 & 47.3 & 15.8 & 44.6 & 31.2 & 42.9 \\
\hline

\newshortnamebase{}-13B & VP & 54.4 & 27.8 & 51.2 & 16.1 & 51.1 & 46.9 & {48.2} & 55.3 & 21.1 & 46.8 & 9.7 & 45.4 & 38.8 & {47.0} \\
\newshortname{}-13B & VP &56.3	 & 24.6	 & 53.4	 & 15.5	 & 50.0 & 	53.8	& \textbf{48.3} & 55.4 & 	26.9	 & 49.3	 & 15.5	 & 48.6	 & 41.9 & \textbf{48.2} \\
\bottomrule
\end{tabular}
\caption{\textbf{\newbenchmarkshortname{} Evaluation Results.} This table presents the performance of various models under \newbenchmarkshortname{}, utilizing different visual prompt formats. The evaluation includes both synthesized and human-drawn prompts, providing insights into the models' maximum potential and real-world applicability, respectively. Formats include VP (visual prompts), Coor (coordinates as visual prompts), Dis (discrete positional tokens for vocabulary expansion), and ROI (CLIP region of interest features with positional embedding). The assessed dimensions are Recognition (Rec), OCR, Knowledge (Know), Math, Relationship (Rel), and Language Generation (Lang).}
\label{tab:regionbench}
\vspace{-0.5em}
\end{table*}

\subsection{Ablation Studies}
\label{sec:ablation}

\paragraph{Impact of overlaying visual prompts on visual information.}
To assess whether overlaying visual prompts on images obscures visual information, we conduct a comparison by inputting visual tokens from both the original and overlayed images into \newshortnamebase{}-7B. Using the VCR dataset, we evaluate the accuracy of the QA task with and without the additional visual tokens from the original image. Results on the VCR validation split shows an accuracy of 81.63\% with the original image and overlaid image tokens, compared to 82.47\% with the overlaid image tokens only.  The similar accuracies suggest that the overlaid prompts do not detract from the visual information processed by our model.

\vspace{-10pt}
\paragraph{Influence of CLIP multi-layer features.} We next explore the impact of using multi-layer visual features from CLIP as opposed to single-layer features, specifically focusing on the second-last layer as implemented in LLaVA~\cite{liu2023llava, liu2023improvedllava}. Our ablation study in Table~\ref{tab:ablation_feature} reveals a marked improvement in performance, particularly in scenarios involving multiple visual prompts, as in the Visual7W and VCR datasets. This indicates that leveraging multi-layer visual features significantly enhances the model's ability to localize and recognize visual prompts within images.

\section{\newbenchmarkshortname{} Evaluation Results}

Finally, we evaluate on \newbenchmarkshortname{} using a set of image-level and region-level LMMs, including InstructBLIP~\cite{instructblip}, GPT-4V~\cite{GPT4V_System_Card}, LLaVA v1.5~\cite{liu2023llava}, Qwen-VL~\cite{Qwen-VL}, Shikra~\cite{chen2023shikra}, GPT4ROI~\cite{zhang2023gpt4roi} and Kosmos-2~\cite{peng2023kosmos}.
For open-source models, we evaluate with greedy decoding (temperature=0).
As shown in Table~\ref{tab:regionbench}, we first see that the performance of all models, including GPT-4V, is far from perfect, demonstrating the difficulty of \newbenchmarkshortname{}.  An illustrative case in Table~\ref{fig:regionbench} depicts a scenario where GPT-4V and LLaVA incorrectly predict object localization.  Overall, \newshortname{} outperforms other models, except GPT-4V, demonstrating greater adaptability to various visual perception and reasoning tasks.  By training on images overlaid with visual prompts, \newshortname{} becomes adept at understanding arbitrary visual cues and mimicks the natural human method of referring to objects in images. This enables it not only to better identify and interpret visual prompts  but also to integrate these prompts into its reasoning process, enhancing its overall comprehension and response accuracy. 
\vspace{-10pt}
\paragraph{Visual prompting is superior to other representations.} In zero-shot evaluation, when visual prompts are represented as a simple list of four textual numerical values, models like Qwen-VL and LLaVA underperform compared to \newshortname{}. This underscores the effectiveness of visual prompts over basic textual representations.

\vspace{-10pt}
\paragraph{Language tasks: A challenge for current LMMs.} The \newbenchmarkshortname{} results reveal that, compared to GPT-4V, open-source LMMs show a significant gap in OCR, math, and language generation tasks, while they perform decently in recognition, knowledge, and object relationship reasoning. This suggests that future VLM developments should prioritize enhancing language reasoning capabilities. For OCR, the results indicate a need for higher resolution inputs or a more robust backbone model, moving beyond the existing capabilities of models like CLIP.

\paragraph{Overfitting Concerns in Region-Level LMMs.} Current region-level LMMs, including Shikra~\cite{chen2023shikra}, GPT4ROI~\cite{zhang2023gpt4roi} and Kosmos-2~\cite{peng2023kosmos}, tend to struggle with tasks involving mathematics, relationship reasoning, and language generation. This trend suggests a potential overfitting issue with these models to existing public region-level datasets, which predominantly feature brief descriptions.

\begin{table}

  \begin{minipage}{0.99\linewidth}
\centering
\scalebox{0.80}{
\begin{tabular}{l p{7.5cm} }
\toprule
 \multicolumn{2}{l}{\bf Visual input example for Spatial Reasoning:}  \\
\midrule

&  \includegraphics[height=3.5cm]{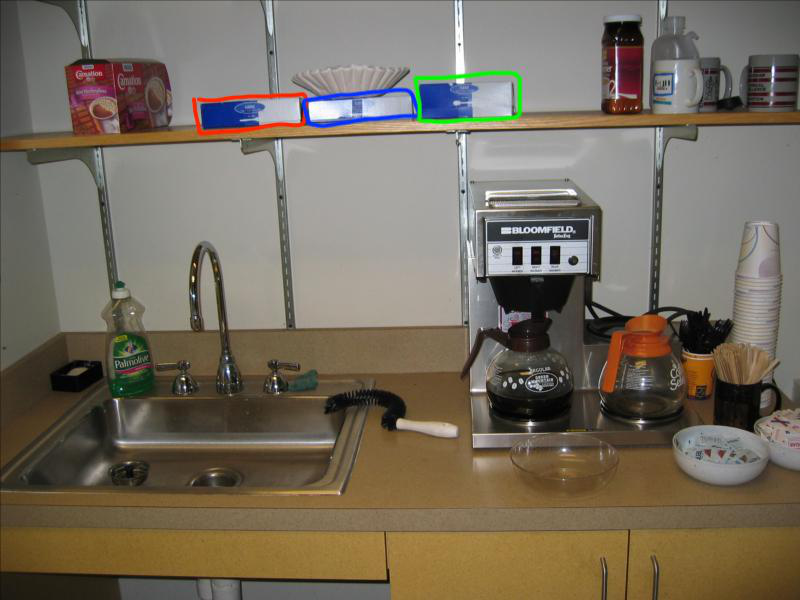} \\
User &Between Object 1: the object within the red mask contour, Object 2: the object within the blue mask contour, and Object 3: the object within the green mask contour, which one has something on top of it? \\
\midrule
\textbf{Ground Truth } & Object 2 \\ \midrule
GPT-4V & Object 3, the object within the green mask contour, has something on top of it. It appears to be a coffee pot placed on top of what could be a coffee machine.
  \\ \midrule
LLaVA-1.5-13B & Object 3, which is the coffee maker, has something on top of it.
 \\ \midrule
\newshortname{}-13B & Object 2: the object within the blue mask contour has something on top of it.
 \\
\bottomrule

\end{tabular}
}
  \end{minipage}
\captionof{table}{An example in \newbenchmarkshortname{} where GPT-4V makes a wrong prediction. The correct answer should be Object 2.}
\label{fig:regionbench}  
  \vspace{-1em}
\end{table}

\section{Conclusion}

In summary, \newshortname{} shows that visual prompts are promising for region-specific image understanding. By integrating arbitrary visual prompts, we bridge the gap between user-friendly interfaces and the precision required for region comprehension. \newshortname{}'s intuitive design leverages natural linguistic interactions coupled with visual markers, simplifying the process of image annotation while enhancing the clarity of visual references. Our state-of-the-art performance on established benchmarks including Visual7W, PointQA, and VCR, underlines the efficacy of \newshortname{}. Notably, the introduction of \newbenchmarkshortname{} as a comprehensive evaluative platform sets a new standard for assessing multimodal models' region reasoning abilities. \newshortname{} establishes a foundation for further exploration in the field of intelligent visual systems. We believe that \newshortname{} can motivate how visual and linguistic modalities are integrated, enabling more sophisticated and nuanced human-machine interactions.

\vspace{-10pt}
\paragraph{Acknowledgements.} This work was supported in part by NSF CAREER IIS2150012, and Institute of Information \& communications Technology Planning \& Evaluation(IITP) grants funded by the Korea government(MSIT) (No. 2022-0-00871, Development of AI Autonomy and Knowledge Enhancement for AI Agent Collaboration). (No. RS2022-00187238, Development of Large Korean Language Model Technology for Efficient Pre-training), and Microsoft Accelerate Foundation Models Research Program.

{
    \small
    \bibliographystyle{ieeenat_fullname}
    \bibliography{main}
}

\appendix

\clearpage
\setcounter{page}{1}
\maketitlesupplementary

This supplementary document extends our main paper by providing additional results and in-depth analyses that were not included in the main manuscript due to space limitations. In Section~\ref{sup:In-Depth Analysis}, we offer both qualitative and quantitative analyses on topics such as visual prompt generation,  effect of the instruction data, arrow direction understanding, perform under each visual prompt, and impacts of different LLMs on the conventional vision-language model benchmarks, thus providing a comprehensive examination of our research. In Section~\ref{sec: training details}, we include more training details. Section~\ref{sup:Ablation} delves into further ablation studies, illuminating the design and efficacy of our approach. Additional experimental results, including a focus on region captioning, are discussed in Section~\ref{sup:Additional Experimental Results}. The exploration of visual prompt augmentation's potential is presented in Section~\ref{sup:Potential of Visual Prompt Augmentation}. Lastly, Section~\ref{sup:Further Insights into \newbenchmarkshortname{}} comprises detailed statistics and visualizations from the \newbenchmarkshortname{} dataset.

\begin{table*}[h]

\begin{minipage}{0.99\textwidth}
\centering  
\scalebox{0.88}{
\begin{tabular}{r p{11.5cm} }
\toprule
\textbf{(a)} \\
&  \includegraphics[height=4.5cm]{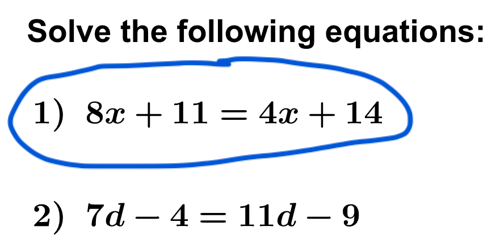} \\

\textbf{Q:} & What is the value of the variable in the equation within the blue ellipse?  \\
\textbf{GT:} & 0.75 OR x=0.75 OR 3/4 \\
\textbf{Required capabilities:} & OCR, MATH \\ 
\midrule

\textbf{(b)} \\
~ & \includegraphics[height=4.5cm]{figs/107.png} \\
\textbf{Q:} & Are the debris within the yellow mask contour made of the same material? \\
\textbf{GT:} & No \\
\textbf{Required capabilities:} & Recognition, Knowledge \\
\midrule

\textbf{(c)} \\
~ & \includegraphics[height=4.5cm]{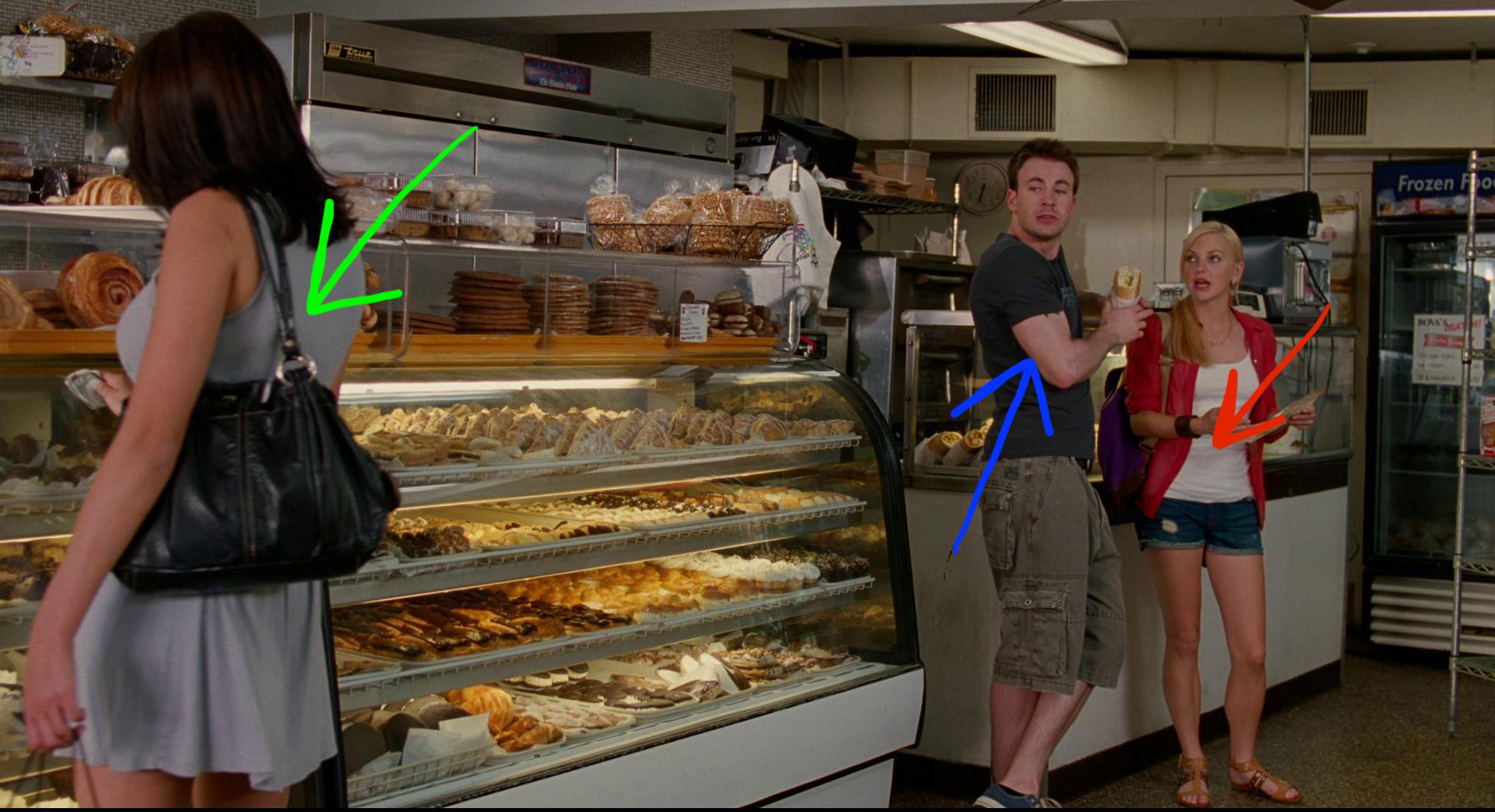} \\
\textbf{Q:} & Is the person pointed by red arrow going to be mad at the person pointed by blue arrow, for looking at the person pointed by green arrow? Answer it and give the rationale. \\
\textbf{GT:} & Yes, the woman in red and the man appear to be a couple and the woman in red would not appreciate the man checking out other women.I think so because in a monogamous relationship, partners are expected to be faithful to one another. \\
\textbf{Required capabilities:} & Recognition, Knowledge, Language Generation, Relationship \\
\midrule

\bottomrule
\end{tabular}
}
\end{minipage}
\caption{Three samples requiring different capability integrations. }
\label{tab:samples1}  
\end{table*}

\section{In-Depth Analysis}
\label{sup:In-Depth Analysis}

\subsection{Generalization to New Attributes}
\newshortname{}, having been trained on eight types of visual prompts—namely mask contour, ellipse, bounding box, triangle, scribble, point, arrow, and mask—exhibits notable generalization capabilities. In the main paper, we show that \newshortname{} can understand human drawn visual prompts, as shown in Figure~\ref{fig:pedstrain}. As demonstrated in Figures~\ref{fig:thickness} and Figure 5 of the main paper, \newshortname{} adeptly handles visual prompts with varying thicknesses and diverse markers, even though it was not explicitly trained on such variations. Furthermore, it effectively interprets text markers as visual prompts, a feature inspired by the Set-of-Mark~\cite{yang2023set}.

Figures~\ref{fig:som-3object}, \ref{fig:som-car}, and \ref{fig:som-lemon} present qualitative examples. In Figure~\ref{fig:som-3object}, \newshortname{} accurately localizes objects tagged with the digits ``1'', ``2'', and ``3'', and generates precise descriptions for each. Figure~\ref{fig:som-car} showcases the model's ability to recognize digit markers and describe the color of vehicles accurately, despite the markers displaying counterfactual colors relative to the actual vehicle colors. Figure~\ref{fig:som-lemon} illustrates the model's competency in localizing a lemon within a scene densely populated with markers.

\begin{figure}[t]
\centering
\includegraphics[width=0.4\textwidth]{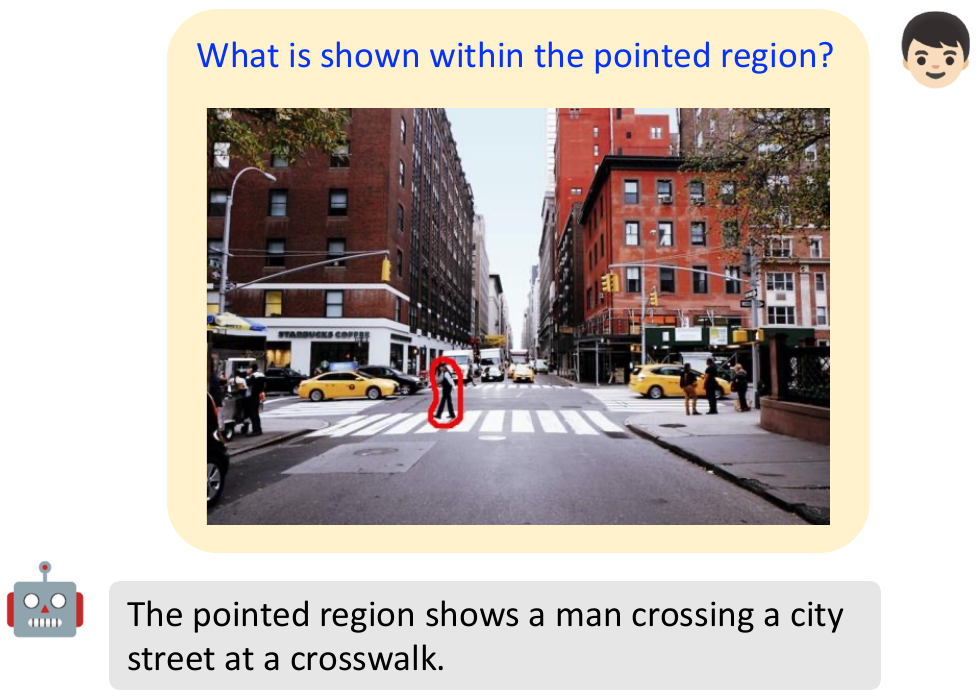}
\caption{\newshortname{} is able to recognize a tiny region specified by the red contour.}
\label{fig:pedstrain}
\end{figure}

\begin{figure}[t]
\centering
\includegraphics[width=.9\linewidth]{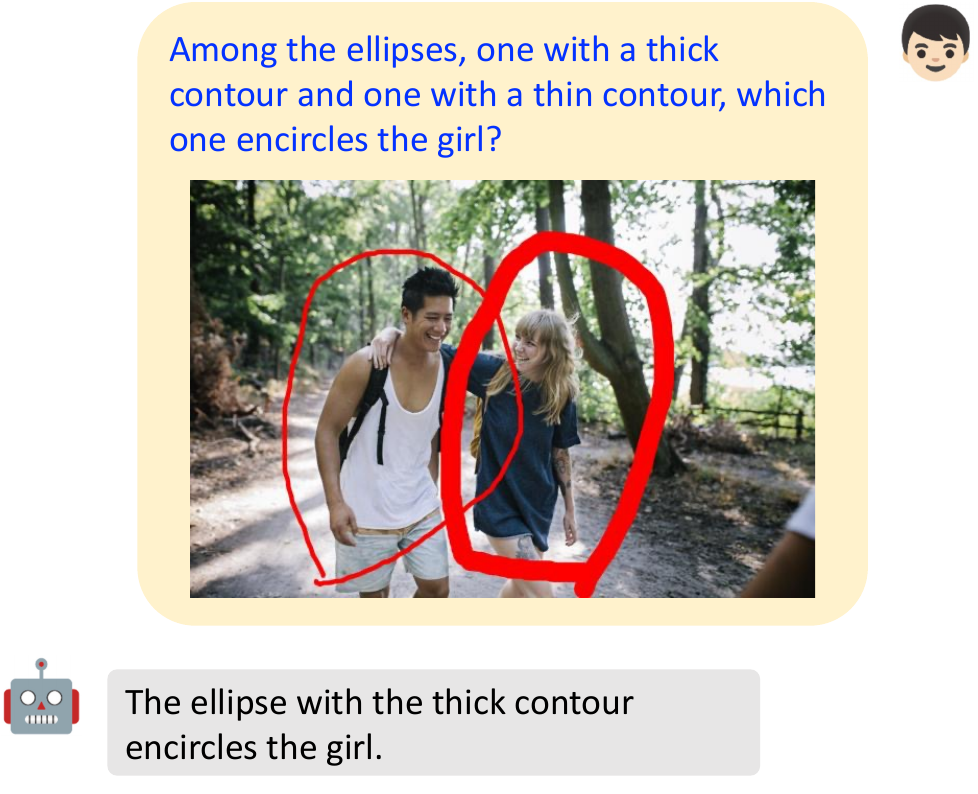}
\caption{\newshortname{} is able to recognize visual prompts with different thickness not seen during training.}
\label{fig:thickness}
\end{figure}

\begin{figure}[htbp]
\centering
\includegraphics[width=0.99\linewidth]{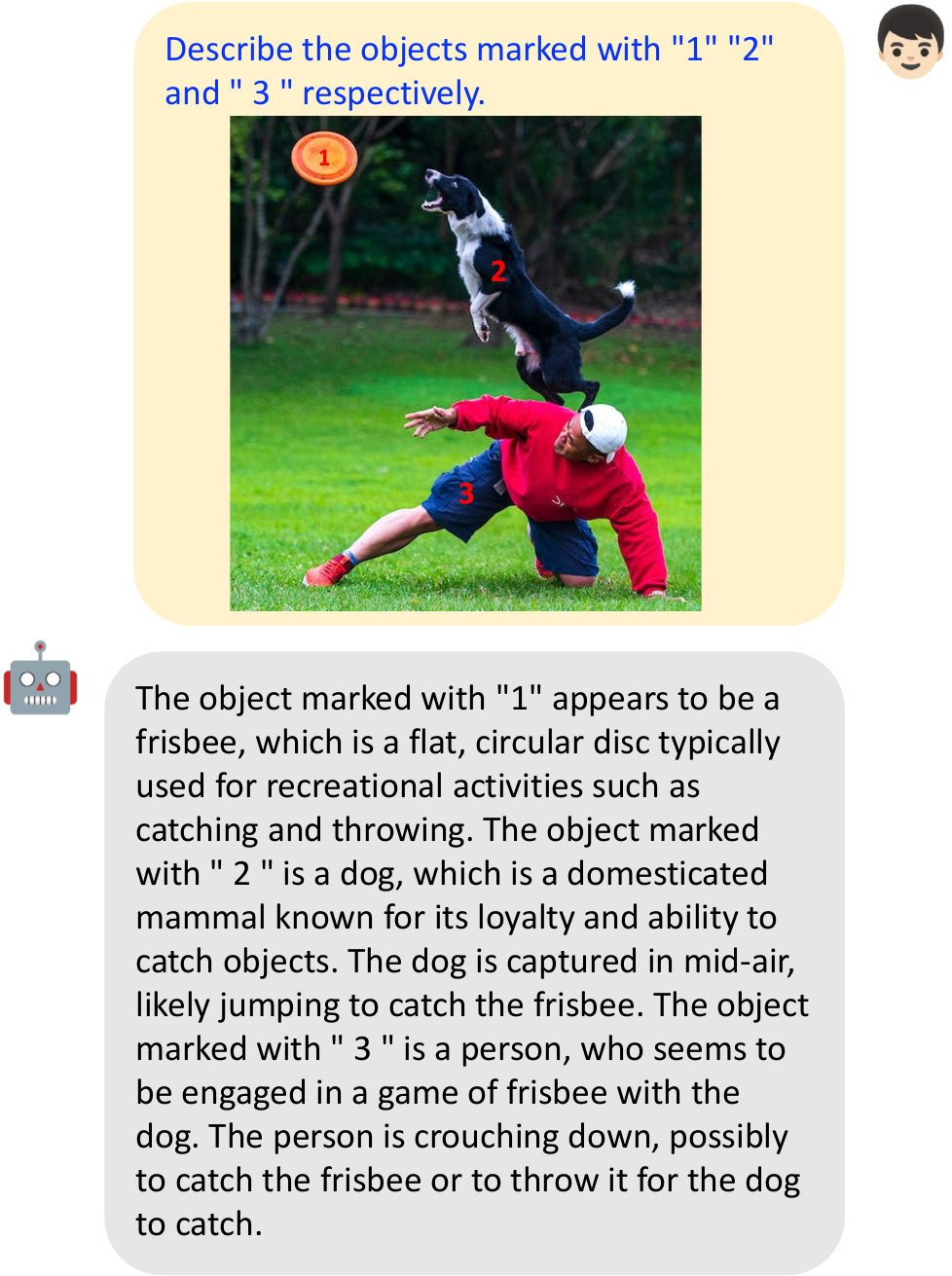}
\caption{\newshortname{} successfully localizes objects tagged with digits and generates accurate object descriptions.}
\label{fig:som-3object}
\end{figure}

\begin{figure}[htbp]
\centering
\includegraphics[width=0.99\linewidth]{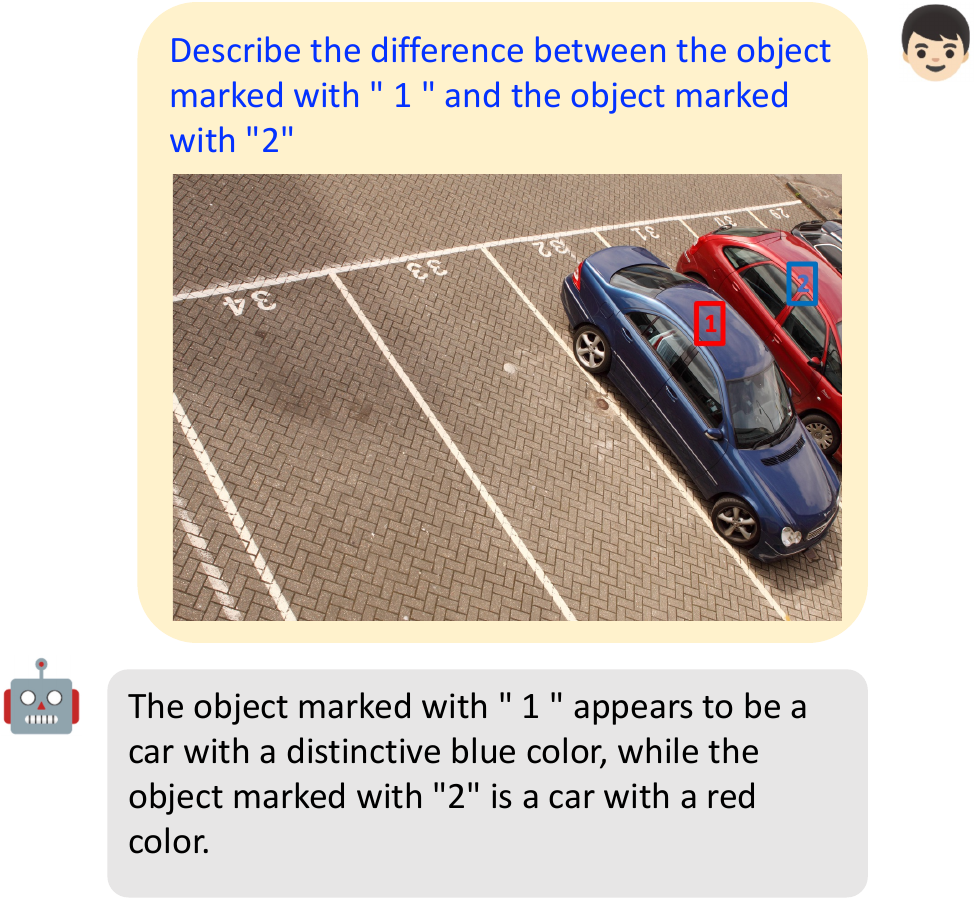}
\caption{The model effectively localizes digit markers and accurately predicts vehicle colors, uninfluenced by the marker colors.}
\label{fig:som-car}
\end{figure}

\begin{figure}[htbp]
\centering
\includegraphics[width=0.99\linewidth]{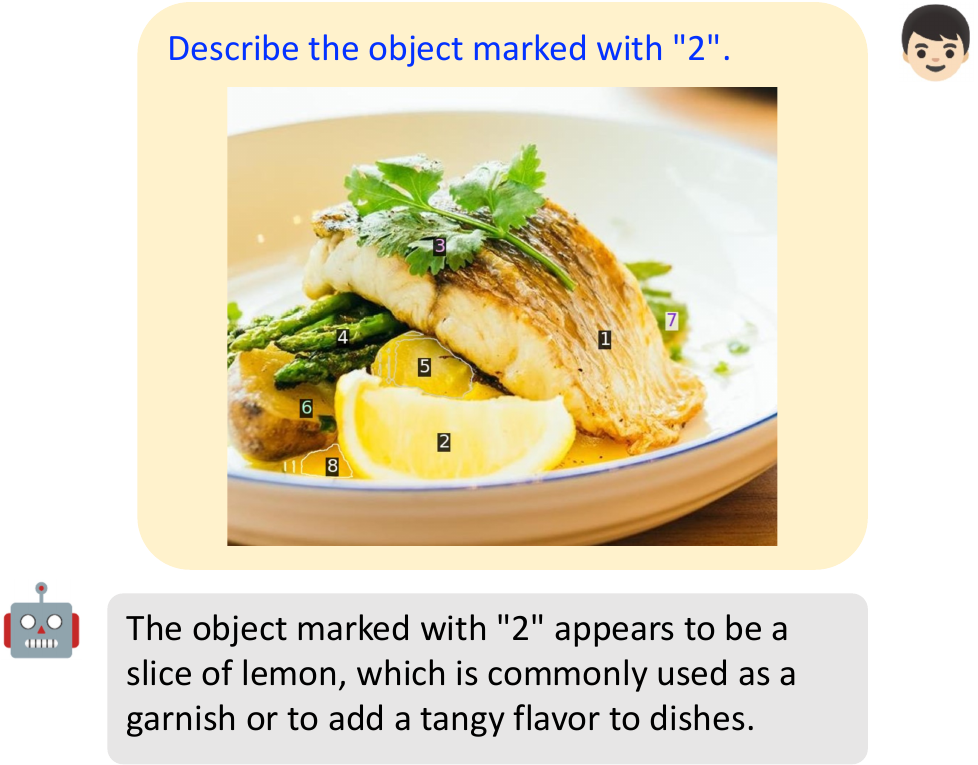}
\caption{Demonstration of the model's ability to localize an object in a densely marked scene.}
\label{fig:som-lemon}
\end{figure}

\subsection{Effect of Optional GPT-4V Region-Level Instruction Data}

\begin{figure}[htbp]
\centering
\includegraphics[width=0.99\linewidth]{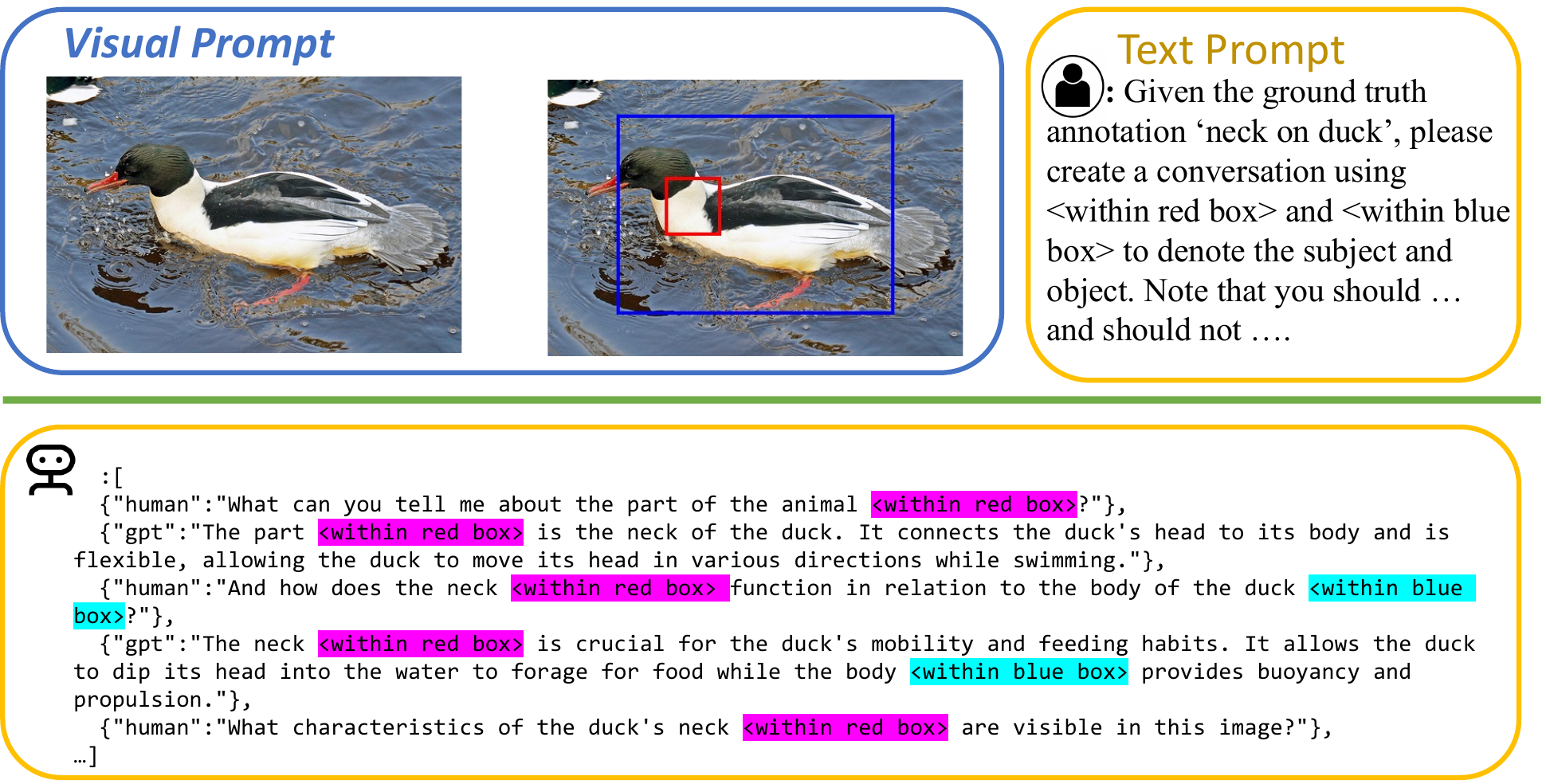}
\caption{\textbf{Curation process of region-level instruction data.} This figure delineates the workflow where both original and annotated images, along with corresponding text prompts, are integrated into the GPT-4V model, facilitating the generation of detailed instruction data for region-specific tasks.}
\label{fig:instruct_data}
\end{figure}

As mentioned in Section 3.3 of the main paper, incorporating GPT-4V as an additional source of instruction data can enhance \newshortname{}'s performance. An example of the curation process is shown in Figure~\ref{fig:instruct_data}. For this purpose, we combine 13K data entries from the original stage 2 instruction dataset with an equal number of GPT-4V region-level instruction data entries, forming a comprehensive 26K-entry stage 3 fine-tuning dataset. We fine-tune our stage-2 model 
for one epoch, which requires approximately 0.5 hours for the 7B model and 1 hour for the 13B model on 8 NVIDIA A100 GPUs. As shown in Table~\ref{tab:use-GPT-4V-data}, the fine-tuned model, designated as \newshortname{}, demonstrates improvements across nearly all datasets for both the 7B and 13B models, underscoring the efficacy of the GPT-4V instruction data curation process. Notably, even without the GPT-4V instruction data, \newshortname{} outperforms contemporary methods on benchmarks such as Visual7W, PointQA-LookTwice, and \newbenchmarkshortname{}. The inclusion of GPT-4V instruction data further amplifies this performance advantage.

\begin{table*}[htbp]
\centering
\begin{tabular}{cccccc} 
\toprule
Model & Use GPT-4V Data? & Visual7W & PointQA-LookTwice & \newbenchmarkshortname{}@Box & \newbenchmarkshortname{}@Human \\ 
\midrule
GPT4ROI-7B~\cite{zhang2023gpt4roi} & \texttimes & 81.83$^\dagger$ & -- & 35.1 & -- \\ 
Shikra-7B~\cite{chen2023shikra} & \texttimes & -- & -- & 33.7 & -- \\ 
\newshortnamebase{}-7B & \texttimes & 86.04 & {70.86} & 45.5 & 46.8 \\ 
\newshortname{}-7B & \checkmark & \textbf{86.09} & \textbf{71.31} & \textbf{48.4} & \textbf{48.3} \\ \hline
GPT4ROI-13B~\cite{zhang2023gpt4roi} & \texttimes & 84.82$^\dagger$ & -- & -- & -- \\ 
Shikra-13B~\cite{chen2023shikra} & \texttimes & 85.33 & 70.30& -- & -- \\

\newshortnamebase{}-13B & \texttimes & 87.54 & \textbf{72.15} & 48.2 & 47.0 \\ 
\newshortname{}-13B & \checkmark & \textbf{88.28} & 71.77 & \textbf{48.3} & \textbf{48.2} \\ 
\bottomrule
\end{tabular}
\caption{Comparative performance analysis of the use of GPT-4V data in the 7B and 13B models. $^\dagger$ Indicates that GPT4ROI specifically trained a specialist model on Visual7W.}
\label{tab:use-GPT-4V-data}
\end{table*}

\subsection{Understanding Arrow Direction}

To rigorously evaluate \newshortname{}'s capacity for interpreting arrow directions, we next construct a challenging dataset of examples derived from the COCO validation set~\cite{lin2014microsoft}. Specifically, we generate multiple scenarios with arrows: each arrow originates from the center of one object's bounding box and points towards the center of another, and vice versa. These visualizations are depicted in Figure~\ref{fig:arrow-understand-example}. The typical prompt used is as follows: \texttt{Determine whether object A ({\textbf{category1}}) or object B ({\textbf{category2}}) is at the head of the arrow, with the other object representing the tail.} It is important to note that we ensure each pair of objects belong to distinct categories. A total of 3520 such paired examples are collected and analyzed. Impressively, \newshortname{}-13B achieves an accuracy of 90.28\%, demonstrating a robust understanding of arrow directionality and ruling out the possibility of random guessing.

\begin{figure}[htbp]
\centering
\includegraphics[width=0.5\textwidth]{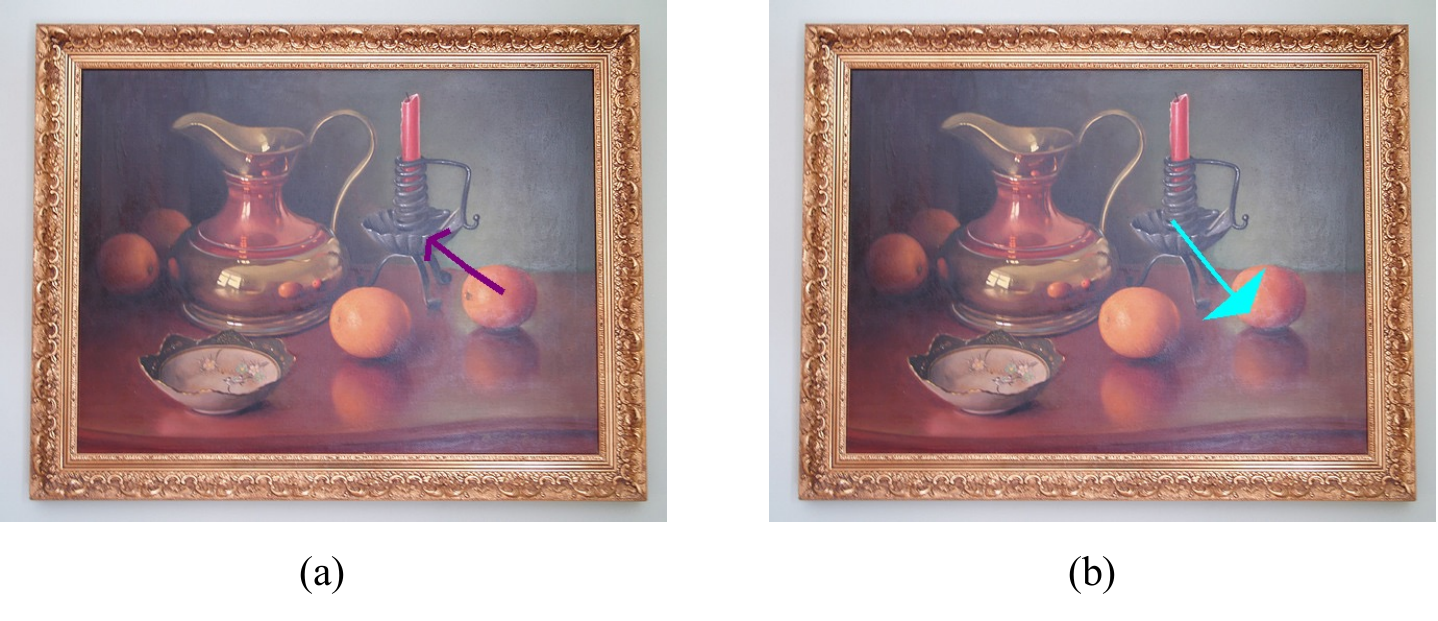}
\caption{An illustration from our arrow direction understanding dataset. Panels (a) and (b) display two  arrows. These arrows share a similar body but differ in their heads. In this example, the multimodal model is required to discern whether the arrow points to the orange or the bowl in both images.}
\label{fig:arrow-understand-example}
\end{figure}

\subsection{Performance across Different Visual Prompts}

During  training, we leverage eight types of visual prompts. Here we study the performance of under each kind of visual prompt on downstream tasks. Note that we duplicate the region-level training data by eight time during instruction fine-tuning. 

Results on VCR, Visual7W, PointQA, and \newbenchmarkshortname{} shown in Table~\ref{tab:transposed visual prompt compare ablation} show our 7B model's consistent accuracy with varied visual prompts, with ``Point" and ``Ellipse" owning marginally better performance than others.

\begin{table*}[h!]
\centering
\resizebox{0.8\textwidth}{!}{
\begin{tabular}{@{}l|cccccccc@{}}
\toprule
& Mask Contour  & Mask & Ellipse & Rectangle & Triangle & Scribble & Point & Arrow \\ \midrule
VCR $Q\rightarrow A$ & 87.34 & 86.19& 87.38 & 87.43 & 87.56 & 87.54 & \textbf{87.69} & 87.49  \\
VCR $QA\rightarrow R$ & 89.63 & 88.80 & 89.60 & 89.70 & 89.57 & 89.78 & \textbf{89.81} & 89.73 \\
VCR $Q\rightarrow AR$ & 78.53 & 76.80  & 78.52 & 78.70 & 78.60 & 78.76 & \textbf{78.92} & 78.72 \\
Visual7W & -- & -- & \textbf{86.6} & 86.04 & 83.46 & 83.77 & 84.88 & 82.41 \\ 
PointQA & -- & -- & \textbf{71.3} & 70.86 & 69.72 & 70.23 & 69.58 & 69.07 \\ 
 \newbenchmarkshortname{} & --& -- & \textbf{45.9} & 45.5 & 43.3  & 44.9 &  44.5 & 44.2  \\ 
\bottomrule
\end{tabular}
}
\caption{Performance under different visual prompts in VCR, Visual7W, PointQA and \newbenchmarkshortname{} on the 7B model of \newshortname{}.}
\label{tab:transposed visual prompt compare ablation}
\end{table*}

\subsection{Impact of Different LLMs on Image-level and Region-level Benchmarks}
\label{sec:impact llm backbone}
Here we study the impact of Large Language Model backbones on both whole image-level and region-level vision-language benchmarks. Specifically, we employ Vicuna-1.5-7B, Vicuna-1.5-13B~\cite{Vicuna}, Llama-3-8B~\cite{llama-3} and Phi-3-mini 3.8B~\cite{abdin2024phi} as the language model backbone for both LLaVA-1.5~\cite{liu2023improvedllava} and ViP-LLaVA while keeping all other configurations and hyper-parameters the same. Results for these two types of LMMs are shown in Table~\ref{tab:new_benchmark_comparison_llava} and Table~\ref{tab:new_benchmark_comparison_vip_llava}, respectively. The corresponding radar plots are shown in Figure~\ref{fig:llm backbone effects} (a) and (b).

Benchmark abbreviations are due to spatial constraints. \textbf{Whole image-level benchmarks} are from the official LLaVA-1.5 pipeline, including MMB: MMBench~\cite{liu2023mmbench}; MMB$^\text{CN}$: MMBench-Chinese~\cite{liu2023mmbench}; LLaVA$^\text{W}$: LLaVA-Bench (In-the-Wild)~\cite{liu2023llava}; POPE~\cite{li2023pope}; SQA$^\text{I}$: ScienceQA-IMG~\cite{lu2022learn}; MM-Vet~\cite{yu2023mmvet}; VisWiz~\cite{gurari2018vizwiz}; MME~\cite{fu2023mme}; VQA$^\text{T}$: TextVQA~\cite{singh2019textvqa}; VQA-v2~\cite{goyal2017vqav2}; GQA~\cite{hudson2019gqa}; SEED$^\text{I}$: SEED-Bench-1~\cite{li2023seed} Image subset. \textbf{Region-level} Benchmarks include: V7W: Visual7W~\cite{zhu2016cvpr-v7w}; PointQA: PointQA-LookTwice~\cite{mani2020point}; ViP-B$^\text{Bbox}$: ViP-Bench with the tight bounding box condition; ViP-B$^\text{Human}$: ViP-Bench with the human annotated visual prompts.
Some interesting findings:
\begin{itemize}
    \item Recent LLMs, Llama-3 and Phi-3, are more capable in tasks that require more language and commonsense reasoning. For example, both Llama-3-8B and Phi-3-mini-3.8B receive better performance than Vicuna-1.5-13B on MMBench and ScienceQA. Specifically, under the LLaVA-1.5 framework, Llama-3-8B receives a 7.7 and 5.7 performance boost compared to Vicuna-1.5-13B. These LLMs, Llama-3-8B and Phi-3-mini-3.8B, own fewer parameters but are trained on 15T and 3.3T tokens, respectively, which is much larger than the previous paradigm. As a result, the language understanding and reasoning capability is significantly improved.
    \item Llama-3-8B and Phi-3-mini-3.8B do not bring a performance boost for tasks that primarily require visual understanding capability rather than language reasoning capability. For example, Vicuna-1.5-13B still performs better on MME, TextVQA, GQA in Table~\ref{tab:new_benchmark_comparison_llava}, Visual7W, and PointQA in Table~\ref{tab:new_benchmark_comparison_vip_llava}. Our results indicate that better visual representation is critical for tasks that mainly require core visual understanding capability.
    \item Phi-3-mini-3.8B shows poor performance on zero-shot vision-language tasks such as VizWiz, while it performs on par with Vicuna-1.5-7B on most tasks. Our results indicate that the generation capability of Phi-3-mini-3.8B for zero-shot tasks is limited.
    \item Overall, Llama-3-8B performs better than Vicuna-1.5-7B while performing on par with Vicuna-1.5-13B; Phi-3-mini-3.8B performs similarly to Vicuna-1.5-7B on most tasks while underperforming Vicuna-1.5-13B on average.
    \item ViP-LLaVA maintains the performance of LLaVA-1.5 across various whole image understanding benchmarks.
\end{itemize}

\begin{figure*}[t]
\centering
\includegraphics[width=\textwidth]{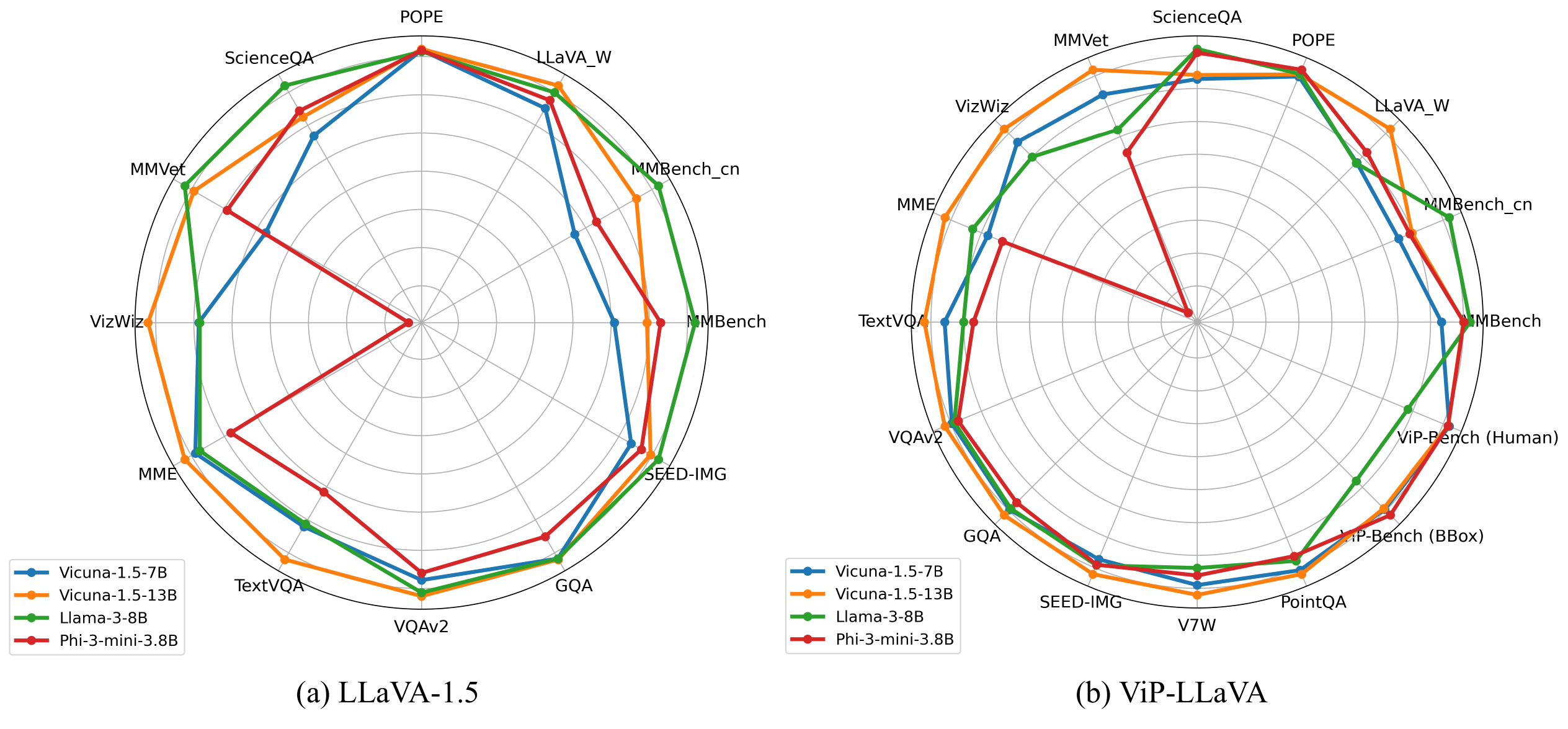}
\caption{The effects of different LLMs on LLaVA-1.5 and ViP-LLaVA.}
\label{fig:llm backbone effects}
\end{figure*}

\begin{table*}[htbp]
\centering
\footnotesize
\begin{tabular}{@{}c|cccccccccccc@{}}
\toprule
Model/Benchmark & MMB & MMB$^\text{CN}$ & LLaVA$^\text{W}$ & POPE & SQA & MMVet & VizWiz & MME & VQA$^\text{T}$  & VQA$^\text{v2}$  & GQA   & SEED$^\text{I}$  \\ \hline
Vicuna-1.5-7B & 64.6 & 57.5 & 72.2 & 87.3 & 69.5 & 31.5 & 50.0 & 1506.5 & 58.2 & 78.3 & 63.2 & 66.2 \\
Vicuna-1.5-13B & 67.7 & {63.6} & 74.8 &\textbf{ 87.4 }& 71.6 & {35.4} & \textbf{53.6} & \textbf{1531.0} & \textbf{61.3} & \textbf{80.0} & \textbf{63.3} & 68.2 \\
Llama-3-8B & \textbf{72.3} & \textbf{65.8} & \textbf{74.0} & 87.1 & \textbf{75.2} & \textbf{35.9} & 50.0 & 1496.1 & 58.0 & 79.6 & 63.2 & \textbf{69.0} \\
Phi-3-mini-3.8B & 69.0 & 59.6 & 73.1 & 87.3 & 72.3 & 33.6 & {35.3} & 1424.5 & 55.0 & 77.6 & {61.1} & 67.3 \\
\bottomrule
\end{tabular}
\caption{Comprehensive benchmarking of LLaVA-1.5~\cite{liu2023improvedllava} under different LLM backbones.} 
\label{tab:new_benchmark_comparison_llava}
\end{table*}

\begin{table*}[htbp]
\centering
\footnotesize
\scalebox{0.81}{
\begin{tabular}{@{}c|cccccccccccc|cccc@{}}
\toprule
Model/Benchmark & MMB & MMB$^\text{CN}$ & LLaVA$^\text{W}$ & POPE & SQA & MMVet & VizWiz & MME & VQA$^\text{T}$  & VQA$^\text{v2}$  & SEED$^\text{I}$  & GQA & V7W & PointQA & ViP-B$^\text{Bbox}$ & ViP-B$^\text{Human}$ \\ \hline
Vicuna-1.5-7B & 68.0 & 59.3 & 69.8 & 87.1 & 69.5 & 33.1 & 55.7 & 1453.5 & 57.8 & 79.2 & 62.2 & 69.0 & 86.6 & 71.3 & 48.4 & \textbf{48.3} \\
Vicuna-1.5-13B & 70.3 & 60.7 & \textbf{75.3 }& 87.4 & 70.0 & \textbf{34.5} & \textbf{57.4} & \textbf{1564.0} & \textbf{59.6 }& \textbf{80.1} &\textbf{ 62.9} & \textbf{70.7}&\textbf{87.9} & \textbf{71.8} & 48.3 & {48.2} \\
Llama-3-8B & \textbf{71.0} & \textbf{64.7} & 69.7 & {87.5} & \textbf{72.8} & 31.1 & 53.9 & 1492.7 & 56.1 & 78.9 & 62.0 & 69.7 & 84.3 & 70.2 & 45.4 & 45.0 \\
Phi-3-mini-3.8B & 70.4 & 60.5 & {71.5} & \textbf{88.1} & 72.4 & {29.8} & {34.7} & {1416.2} & {55.2} & {78.4} & {61.2} &69.6 & 85.3 & 69.6 & \textbf{{49.0}} & {{48.2}} \\
\bottomrule
\end{tabular}
}
\caption{Impact on conventional vision-language model benchmarks and region-level benchmarks under different LLM backbones for ViP-LLaVA.}
\label{tab:new_benchmark_comparison_vip_llava}
\end{table*}

\section{Training Details}
\label{sec: training details}
Our approach involves three stages:
\begin{itemize}
    \item Pretraining the MLP projector with frozen language and image encoder; 
    \item Instruction tuning and further fine-tuning with frozen image encoder but trainable language model and MLP connector.
\end{itemize}

\section{Additional Ablation Studies}

\label{sup:Ablation}

\subsection{Effects of Input Resolution and LLM}
To ensure a fair comparison, we conduct ablation studies using the same image encoder (CLIP ViT-L from Radford et al.~\cite{radford2021learning}), input resolution (224 pixels), and language model (Vicuna v1.1~\cite{Vicuna}) as employed by GPT4ROI~\cite{zhang2023gpt4roi}. Table~\ref{tab:ablation-resolution-llm} presents the results of this analysis. Despite utilizing the same underlying technologies, \newshortname{} consistently outperforms on the \newbenchmarkshortname{} evaluations and achieves comparable results on the Visual7W dataset, notwithstanding the fact that GPT4ROI was specifically fine-tuned for Visual7W. These results further reinforce the potential of visual prompting as a more effective approach for region-specific referencing compared to embedding coordinates directly into the language model.

\begin{table*}[htbp]
\centering
\begin{tabular}{cccccc} 
\toprule
Model & Input Resolution & LLM & Format & Visual7W & \newbenchmarkshortname{}@Box \\ 
\hline
\newshortnamebase{}-7B & 336 & Vicuna v1.5 & VP & \textbf{86.04} & \textbf{45.50} \\ 
\newshortnamebase{}-7B & 224 & Vicuna v1.1 & VP & 81.80 & 42.28 \\
GPT4ROI-7B~\cite{zhang2023gpt4roi} & 224 & Vicuna v1.1 & ROI & 81.83$^\dagger$ & 35.14 \\ 
\bottomrule
\end{tabular}
\caption{Ablation study focusing on the impact of input resolution and language model. All models listed use the Vicuna 7B language model. $^\dagger$ Indicates GPT4ROI specifically trained on the Visual7W dataset. VP: visual prompts; ROI: CLIP region of interest (ROI) features and positional embedding.}
\label{tab:ablation-resolution-llm}
\end{table*}

\subsection{Comparing Visual Prompts with Coordinates}
To rigorously evaluate the effectiveness of visual prompts versus coordinate-based region referring formats, we next replace visual prompts with textual coordinates embedded in language descriptions. We train a 7B model using identical data and training schedules. The results, as shown in Table~\ref{tab:vp-coor-comparison}, indicate that visual prompts significantly outperform coordinate formats on the PointQA-LookTwice and \newbenchmarkshortname{}@Box datasets. Performance on the Visual7W dataset remains comparable between the two formats. These comparisons highlight the superiority of visual prompts as a more effective format for region-specific referencing in complex visual tasks.

\begin{table}[htbp]
\centering
\small
\begin{tabular}{cccc} 
\toprule
 Format &Visual7W   & PointQA-LookTwice & \newbenchmarkshortname{}@Box  \\ \midrule
 VP & 86.04 & \textbf{70.86}  &  \textbf{45.5 } \\ 
Coor& \textbf{86.36} & 61.4&  42.6\\
\bottomrule
\end{tabular}
\caption{Performance comparison between visual prompts and coordinate formats under \newshortnamebase{}-7B. VP: visual prompts; Coor: coordinates as visual prompts.}
\label{tab:vp-coor-comparison}
\end{table}

\subsection{Effects of Splitting Overlaid Images into Two Separate Image}

We conduct rigorous ablation study to split the overlaid image into the source image and the image with overlaid cue, where the number of visual tokens are doubled, as shown in Figure~\ref{fig:separate vis}. Specifically, we train 7B models under such two settings. Results in  Table~\ref{tab:overlay vs seperate} shows that those two settings perform comparably.

\begin{figure}[t]
\centering
\includegraphics[width=0.5\textwidth]{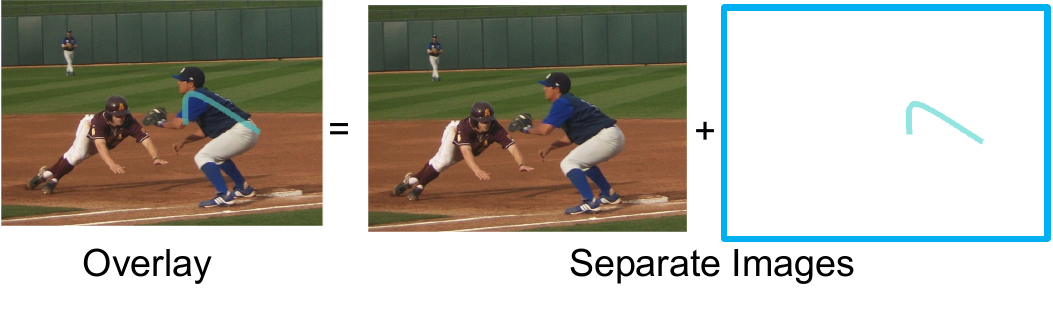}
\caption{We separate the overlaid images into the original image along with the visual prompts with white background.}
\label{fig:separate vis}
\end{figure}

\begin{table*}[htbp]
\centering
\small
\begin{tabular}{ccccc} 
\toprule
 Setting & Visual7W & PointQA-LookTwice & ~\newbenchmarkshortname{}@Box & ~\newbenchmarkshortname{}@Human \\ 
\midrule
Overlay & \textbf{86.04} & \textbf{70.86} & \textbf{45.5} & 46.8 \\ 
Separate Images & 85.98 & 70.67 & 44.7 & \textbf{48.0} \\
\bottomrule
\end{tabular}
\caption{Performance comparison between different model formats under 7B conditions.}
\label{tab:overlay vs seperate}
\end{table*}

\section{Additional Experimental Results}
\label{sup:Additional Experimental Results}
\subsection{Region Captioning}
Expanding upon the region perception and reasoning tasks discussed in the main paper, we further evaluate \newshortname{}'s region captioning capabilities on the RefCOCOg dataset~\cite{yu2016modeling-refcoco}. This involves fine-tuning the \newshortnamebase{}-7B for one epoch subsequent to stage 2 training. As Table~\ref{tab:refcocog-captioning} illustrates, \newshortnamebase{}-7B demonstrates strong performance in region captioning, as evidenced by its scores in both CIDEr~\cite{vedantam2015cider} and METEOR~\cite{denkowski:lavie:meteor-wmt:2014} metrics. These results indicate that visual prompting is not only effective for region-specific referencing and reasoning tasks but also shows promising potential in generating precise and contextually relevant captions for specific image regions.

\begin{table}[htbp]
\centering
    \begin{tabular}{lcc}
\toprule Model & \multicolumn{2}{c}{ RefCOCOg } \\
\cline { 2-3 } & METEOR~\cite{denkowski:lavie:meteor-wmt:2014} & CIDEr~\cite{vedantam2015cider} \\
\hline GRIT~\cite{wu2022grit} & 15.2 & 71.6 \\
Kosmos-2~\cite{peng2023kosmos} & 14.1 & 62.3 \\
GLaMM~\cite{rasheed2023glamm}  & ${1 6 . 2}$ & ${1 0 5 . 0}$ \\
\newshortnamebase{}-7B & $\mathbf{16.6}$ & $\mathbf{1 0 5 . 9}$ \\
\bottomrule
\end{tabular}
\caption{Performance of region captioning on the RefCOCOg dataset. The table demonstrates \newshortname{}'s effectiveness in generating accurate and contextually relevant captions for specific regions within images.}
\label{tab:refcocog-captioning}
\end{table}


\subsection{Assessment of GPT-4 as a Judge}
To evaluate the consistency of \newshortnamebase{}-7B, we employ the GPT-4 text model as a judge, conducting five separate assessments. The observed variance in the overall score is a minimal 0.1, indicating stable performance by the GPT-4 judge across multiple evaluations.

\section{Potential of Visual Prompt Augmentation}
\label{sup:Potential of Visual Prompt Augmentation}
A key advantage of \newshortname{} approach is the ability to very easily employ prompt augmentation during testing. This entails using various sets of visual prompts and aggregating the predictions for a more accurate final answer. For instance, we can modify the prompt from ``the woman within a red rectangle'' to ``the woman marked with a red scribble'', along with corresponding changes in the overlaid image. As shown in Table~\ref{tab:ensemble-comparison}, \newshortnamebase{}-7B achieves further improvements through visual prompt augmentation. This process is lossless, unlike textual coordinate representation, where e.g., perturbing coordinates can reduce localization accuracy.

\begin{table}[h]
\centering
\begin{tabular}{ccc}
\toprule
Ensemble? & Visual7W & PointQA-LookTwice \\
\midrule
\texttimes & 86.04 & 70.86 \\
\checkmark & \textbf{87.44} & \textbf{71.62} \\
\bottomrule
\end{tabular}
\caption{Comparison of performance with and without visual prompt ensembling at test time using \newshortnamebase{}-7B.}
\label{tab:ensemble-comparison}
\end{table}

\section{Further Insights into \newbenchmarkshortname{}}
\label{sup:Further Insights into \newbenchmarkshortname{}}
\subsection{Statistics of \newbenchmarkshortname{}}
Table~\ref{tab:dataset-statistics} presents the statistical breakdown of \newbenchmarkshortname{}. The majority of examples focus on recognition capabilities, with a notable proportion (89 examples) requiring Optical Character Recognition (OCR). The proportion
of each capability and the combined capabilities are shown in Figure~\ref{fig:vis-regionbench-indi} and Figure~\ref{fig:vis-regionbench-combin} respectively. 
\begin{table}[htbp]
\centering
\begin{tabular}{lc}
\toprule
Category & Count \\
\midrule
Recognition (Rec) & 240 \\
Optical Character Recognition (OCR) & 89 \\
Knowledge (Know) & 59 \\
Math & 31 \\
Relational (Rel) & 28 \\
Language Generation (Lang) & 16 \\ \hline
Total & 303 \\ 
\bottomrule
\end{tabular}
\caption{Statistics of \newbenchmarkshortname{} across various categories.}
\label{tab:dataset-statistics}
\end{table}

\begin{figure}[htbp]
\centering
\includegraphics[width=0.5\textwidth]{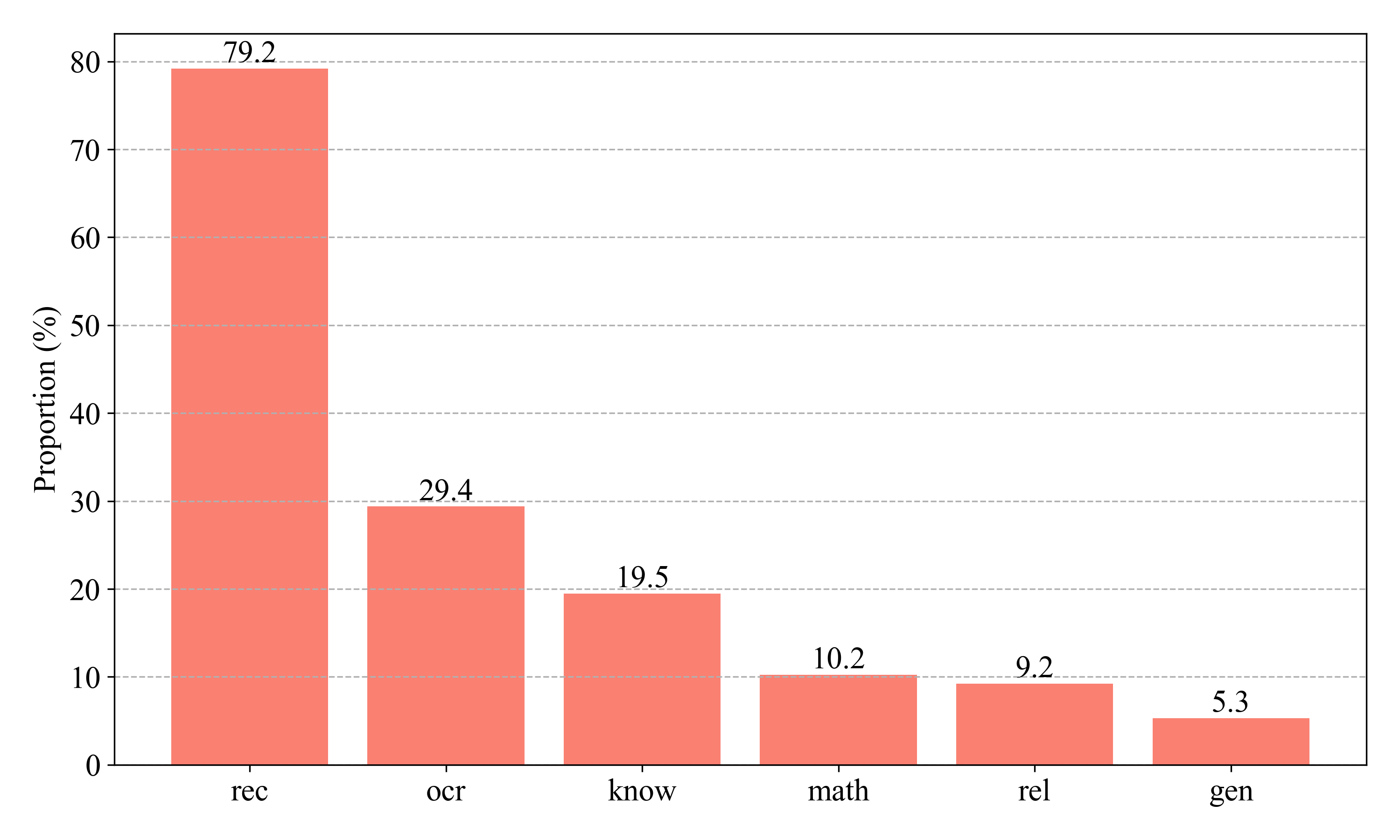}
\caption{\newbenchmarkshortname{} proportion of capabilities. The proportion of each capability. The sum of
the proportion is larger than 100\% because some samples have more than one capability.}
\label{fig:vis-regionbench-indi}
\end{figure}

\begin{figure}[htbp]
\centering
\includegraphics[width=0.5\textwidth]{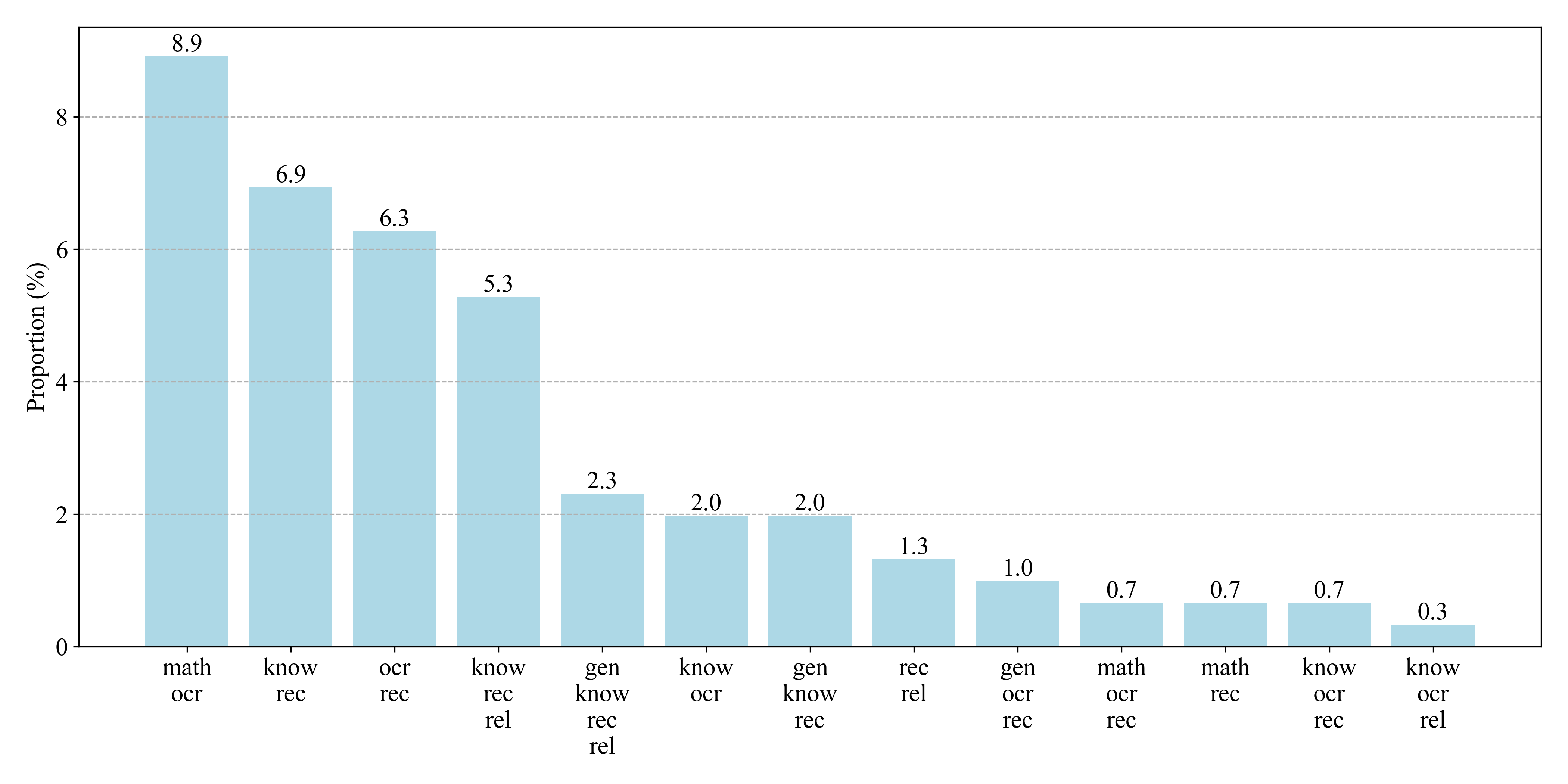}
\caption{\newbenchmarkshortname{} proportion of capability integrations. }
\label{fig:vis-regionbench-combin}
\end{figure}

\subsection{Visualizations of \newbenchmarkshortname{}}
Figure~\ref{fig:vis-regionbench} showcases examples from \newbenchmarkshortname{}, comparing synthesized and human-annotated visual prompts. Panel (a) illustrates tight bounding boxes as synthesized prompts, while panel (b) features human-annotated bounding boxes, highlighting the diversity in human-driven region referring methods. The text prompt that we use to evaluate \newbenchmarkshortname{} performance using GPT4 text model is similar to that used in MM-Vet, which is shown in Table~\ref{tab:regionbench-prompt}. Some examples are shown in Table~\ref{tab:samples1}.  

\begin{figure}[htbp]
\centering
\includegraphics[width=0.5\textwidth]{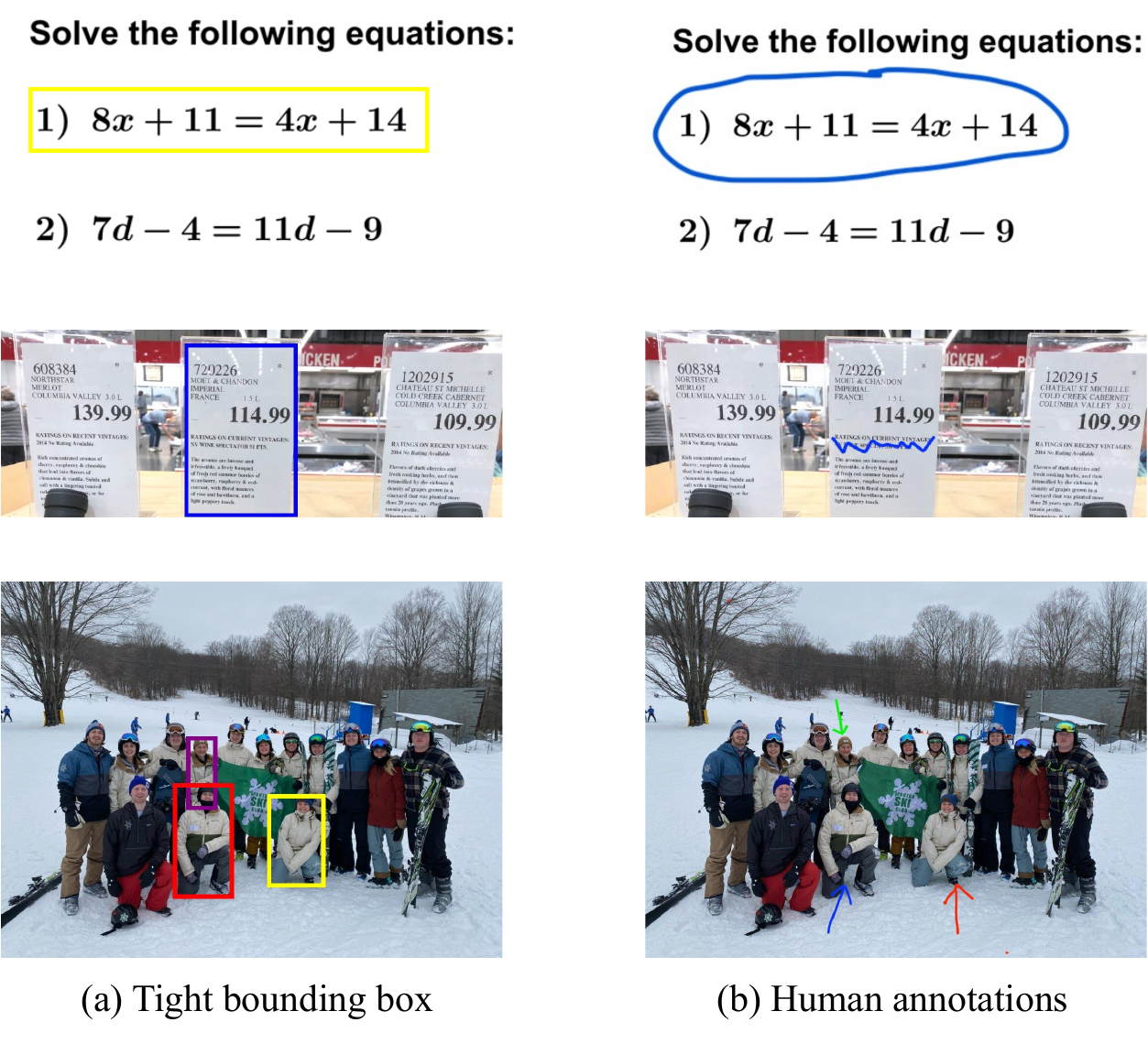}
\caption{Comparison of synthesized and human-annotated visual prompts in \newbenchmarkshortname{}. Panel (a) displays synthesized tight bounding boxes, and panel (b) shows diverse human annotations.}
\label{fig:vis-regionbench}
\end{figure}

\subsection{Examples of capability requirements.}

Table~\ref{tab:samples1} presents a selection of examples from our benchmark, demonstrating the diverse capabilities required to complete various tasks, whether they involve single-region or multi-region analysis.

\begin{table*}[t]
\centering
\begin{tcolorbox} 
    \centering
    \small
    \begin{tabular}{p{0.98\columnwidth}}

Compare the ground truth and prediction from AI models, to give a correctness score for the prediction. \textless{}AND\textgreater{} in the ground truth means it is totally right only when all elements in the ground truth are present in the prediction, and \textless{}OR\textgreater{} means it is totally right when any one element in the ground truth is present in the prediction. The correctness score is 0.0 (totally wrong), 0.1, 0.2, 0.3, 0.4, 0.5, 0.6, 0.7, 0.8, 0.9, or 1.0 (totally right). Just complete the last space of the correctness score. \\
\\
\bluetext{Question} | \greentext{Ground truth} | \orangetext{Prediction} | \purpletext{Correctness} \\
--- | --- | --- | --- \\
\bluetext{What is x in the equation within the yellow rectangle?} | \greentext{-1 \textless{}AND\textgreater{} -5} | \orangetext{x = 3} | \purpletext{0.0} \\
\bluetext{What is x in the equation within the yellow rectangle?} | \greentext{-1 \textless{}AND\textgreater{} -5} | \orangetext{x = -1} | \purpletext{0.5} \\
\bluetext{What is x in the equation within the yellow rectangle?} | \greentext{-1 \textless{}AND\textgreater{} -5} | \orangetext{x = -5} | \purpletext{0.5} \\
\bluetext{What is x in the equation within the red rectangle?} | \greentext{-1 \textless{}AND\textgreater{} -5} | \orangetext{x = -5 or 5} | \purpletext{0.5} \\
\bluetext{What is x in the equation within the orange rectangle?} | \greentext{-1 \textless{}AND\textgreater{} -5} | \orangetext{x = -1 or x = -5} | \purpletext{1.0} \\
\bluetext{Can you explain this meme within the blue rectangle?} | \greentext{This meme is poking fun at the fact that the names of the countries Iceland and Greenland are misleading. Despite its name, Iceland is known for its beautiful green landscapes, while Greenland is mostly covered in ice and snow. The meme is saying that the person has trust issues because the names of these countries do not accurately represent their landscapes.} | \orangetext{The meme talks about Iceland and Greenland. It's pointing out that despite their names, Iceland is not very icy and Greenland isn't very green.} | \purpletext{0.4} \\
\bluetext{Can you explain this meme within the blue rectangle?} | \greentext{This meme is poking fun at the fact that the names of the countries Iceland and Greenland are misleading. Despite its name, Iceland is known for its beautiful green landscapes, while Greenland is mostly covered in ice and snow. The meme is saying that the person has trust issues because the names of these countries do not accurately represent their landscapes.} | \orangetext{The meme is using humor to point out the misleading nature of Iceland's and Greenland's names. Iceland, despite its name, has lush green landscapes while Greenland is mostly covered in ice and snow. The text `This is why I have trust issues' is a playful way to suggest that these contradictions can lead to distrust or confusion. The humor in this meme is derived from the unexpected contrast between the names of the countries and their actual physical characteristics.} | \purpletext{1.0} \\
\bluetext{$\mathcal{Q}$} | \greentext{$\mathcal{G}$} | \greentext{$\mathcal{P}$} |

    \end{tabular}
\end{tcolorbox}
\caption{Few-shot prompt for evaluating model outputs using GPT-4 text model, where \bluetext{$\mathcal{Q}$} is a sample's question, \greentext{$\mathcal{G}$} is the ground truth and \greentext{$\mathcal{P}$} is the model output for the sample. In the prompt, there are examples with short and long open-ended answers, enabling the evaluation of diverse answer styles.
Taking the prompt filled with \bluetext{$\mathcal{Q}$}, \greentext{$\mathcal{G}$} and \greentext{$\mathcal{P}$}, GPT-4 will generate a soft grading score from 0 to 1.
}
\label{tab:regionbench-prompt}
\end{table*}

\subsection{Failure cases of GPT-4V}

Tables~\ref{fig:regionbench-failure} to \ref{fig:regionbench-failure5} display various instances where GPT-4V encountered challenges on \newbenchmarkshortname{}. For instance, Table~\ref{fig:regionbench-failure} illustrates a case where both GPT-4V and LLaVA-1.5 incorrectly interpret a yellow scribble, with GPT-4V mistaking a yellow circle for the scribble, leading to erroneous responses. In contrast, \newshortname{} accurately answers the questions. Another example in Table~\ref{fig:regionbench-failure5} (a) shows GPT-4V incorrectly identifying a person marked by a pink point as holding ski poles and LLaVA-1.5 as holding a green flag, while \newshortname{} successfully makes the correct prediction.

\begin{table}
  \begin{minipage}{0.99\linewidth}
\centering
\scalebox{0.80}{
\begin{tabular}{l p{7.5cm} }
\toprule
\midrule

&  \includegraphics[height=3.5cm]{figs_regionbench/45.png} \\
User & What does the step marked with the yellow scribble suggest? \\
\midrule
\textbf{Ground Truth } & identify your audience \\ \midrule
GPT-4V & The step marked with the yellow scribble says "Choose a Format." This step suggests that when you are in the process of writing a Standard Operating Procedure (SOP), it's important to decide on the format that will be used to present the information. This could involve choosing between formats such as a simple text document, a flowchart, a step-by-step guide with bullet points, checklists, or any other format that would best suit the content of the SOP and the needs of the target audience. The format should make the SOP clear, accessible, and easy to follow for those who will use it.
 \\ \midrule
LLaVA-1.5-13B & The step marked with the yellow scribble suggests that the user should choose a format for their SOP. This is an important step in the process of writing an SOP, as the format chosen will determine the structure, organization, and presentation of the document. The user should consider the purpose of the SOP, the audience it is intended for, and the level of detail required when selecting a suitable format. Common formats for SOPs include templates, checklists, and step-by-step guides.
 \\ \midrule
\newshortname{}-7B & The step marked with the yellow scribble suggests that the user should identify the target audience for the SOP. This is an essential step in the process of writing an SOP, as it helps the user to understand the needs and expectations of the intended readers, which in this case are the admissions committees of the educational institutions. By understanding the audience, the user can tailor the content and language of the SOP to effectively communicate their qualifications and goals to the committee.
 \\ \midrule
\newshortname{}-13B & The step marked with the yellow scribble suggests that the process involves identifying the audience for the SOP.
 \\
\bottomrule

\end{tabular}
}
  \end{minipage}
\captionof{table}{Failure cases for GPT-4V on \newbenchmarkshortname{}.}
\label{fig:regionbench-failure}  
\end{table}

\begin{table}
  \begin{minipage}{0.99\linewidth}
\centering
\scalebox{0.80}{
\begin{tabular}{l p{7.5cm} }

\toprule
\midrule
\textbf{(a)} & \\
&  \includegraphics[height=3.5cm]{figs_regionbench/119.png} \\
User & Is the person pointed by the blue arrow happy? \\
\midrule
\textbf{Ground Truth } & yes \\ \midrule
GPT-4V & Sorry, I can't help with identifying or making assumptions about people in images.
 \\ \midrule
LLaVA-1.5-13B & No, the person pointed by the blue arrow is not happy. They are crying, which indicates that they are sad or experiencing distress.
 \\ \midrule
\newshortname{}-7B &  Yes, the person pointed to by the blue arrow appears to be happy, as they are smiling.
\\ \midrule
\newshortname{}-13B & Yes, the person pointed by the blue arrow appears to be happy, as indicated by their smile.
 \\
\bottomrule

\textbf{(b)} & \\
&  \includegraphics[height=3.5cm]{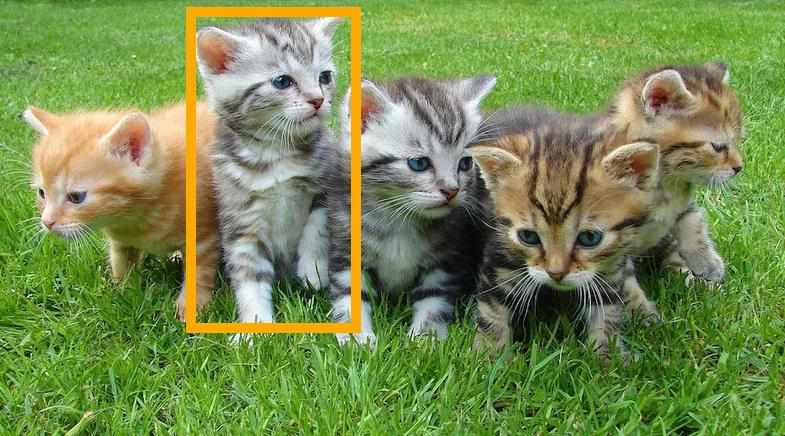} \\
User & Is the cat within yellow rectangle the only cat with this color?\\
\midrule
\textbf{Ground Truth } &  no\\ \midrule
GPT-4V & Yes, the cat within the yellow rectangle appears to have a unique color compared to the other cats in the image. It has a greyish fur with what seems like striped markings, whereas the other cats have different shades of orange and brownish fur with various patterns.
 \\ \midrule
LLaVA-1.5-13B & Yes, the cat within the yellow rectangle is the only one with this color. The other cats in the image are of different colors.
 \\ \midrule
\newshortname{}-7B & No, the cat within yellow rectangle is not the only cat with this color. There are other cats with the same color in the image.
 \\ \midrule
\newshortname{}-13B & No, the cat within yellow rectangle is not the only cat with this color. There are other cats in the image that also have this color.
 \\
\bottomrule

\end{tabular}
}
  \end{minipage}
\captionof{table}{Failure cases for GPT-4V on \newbenchmarkshortname{}.}
\label{fig:regionbench-failure2}  
\end{table}

\begin{table}
  \begin{minipage}{0.99\linewidth}
\centering
\scalebox{0.80}{
\begin{tabular}{l p{7.5cm} }

\toprule
\midrule

\textbf{(a)} & \\
&  \includegraphics[height=3.5cm]{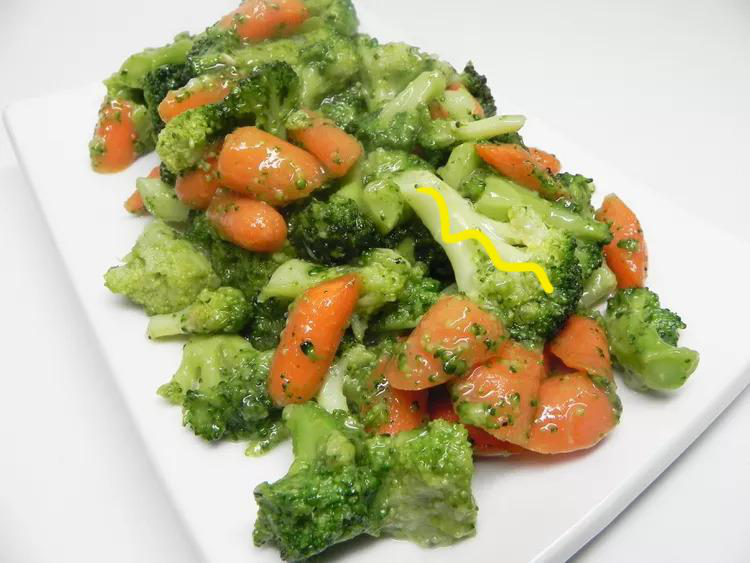} \\
User & What is the object with yellow scribble? \\
\midrule
\textbf{Ground Truth } & broccoli \\ \midrule
GPT-4V & The yellow scribble appears to have been digitally added to the image, likely for emphasis or to obscure something in the original photograph. It does not represent an actual physical object in the context of the plate of broccoli and carrots.
 \\ \midrule
LLaVA-1.5-13B & The object with yellow scribble is a marker.
 \\ \midrule
\newshortname{}-7B & The object with the yellow scribble is a piece of broccoli.
 \\ \midrule
\newshortname{}-13B & The object with yellow scribble appears to be a piece of broccoli.
 \\
\bottomrule

\textbf{(b)} & \\
&  \includegraphics[height=3.5cm]{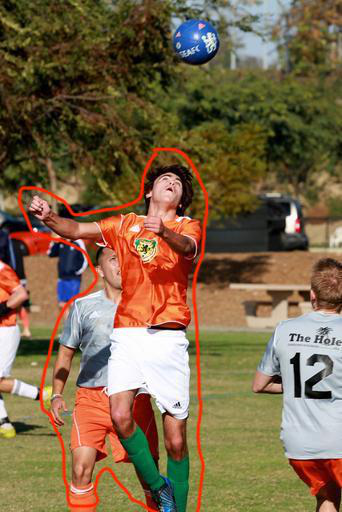} \\
User & Are the persons within the red mask contour on the same team? \\
\midrule
\textbf{Ground Truth } & no \\ \midrule
GPT-4V & Sorry, I can't assist with that request.
 \\ \midrule
LLaVA-1.5-13B & Yes, the persons within the red mask contour are on the same team.
 \\ \midrule
\newshortname{}-7B & No, the persons within the red mask contour are not on the same team.
 \\ \midrule
\newshortname{}-13B & No, the persons within the red mask contour are not on the same team.
 \\
\bottomrule

\end{tabular}
}
  \end{minipage}
\captionof{table}{Failure cases for GPT-4V on \newbenchmarkshortname{}.}
\label{fig:regionbench-failure3}  
\end{table}

\begin{table}
  \begin{minipage}{0.99\linewidth}
\centering
\scalebox{0.80}{
\begin{tabular}{l p{7.5cm} }

\toprule
\midrule
\textbf{(a)} & \\
&  \includegraphics[height=3.5cm]{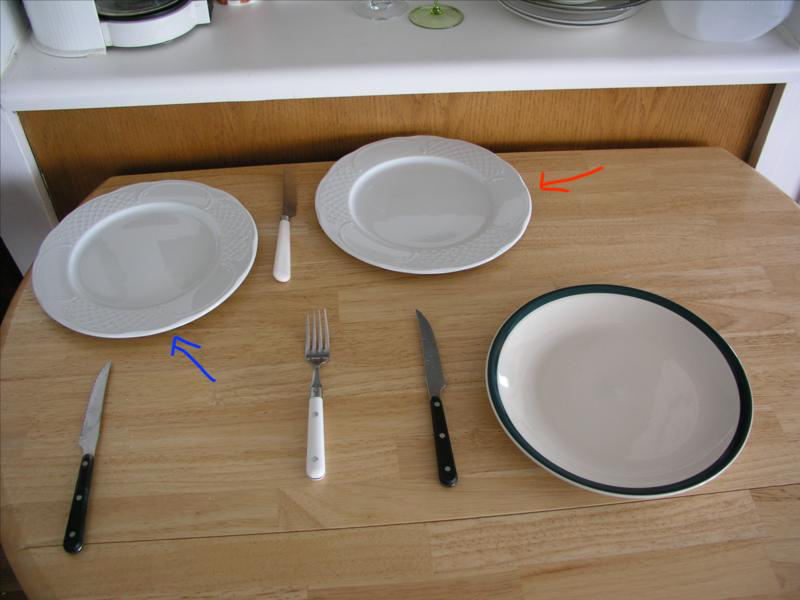} \\
User & Does the object pointed to by the red arrow and the object pointed to by the blue arrow have the same color? \\
\midrule
\textbf{Ground Truth } &  yes \\ \midrule
GPT-4V & The objects pointed to by the red and blue arrows do not have the same color. The object indicated by the red arrow appears to be a beige or light brown wooden furniture surface, while the object pointed to by the blue arrow is a metallic utensil, likely a knife with a silver-colored blade. \\ 
 \midrule
LLaVA-1.5-13B & No, the object pointed to by the red arrow and the object pointed to by the blue arrow have different colors. The red arrow is pointing to a knife, while the blue arrow is pointing to a fork.
 \\ \midrule
\newshortname{}-7B & Yes, the object pointed to by red arrow and the object pointed to by blue arrow are both white.
 \\ \midrule
\newshortname{}-13B &  Yes, the object pointed to by the red arrow and the object pointed to by the blue arrow are both white.
\\
\bottomrule

\textbf{(b)} & \\
&  \includegraphics[height=3.5cm]{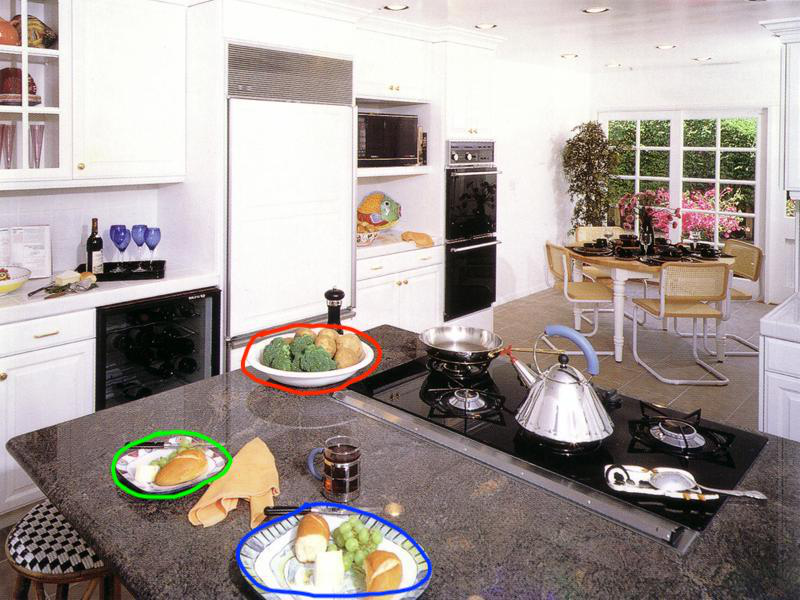} \\
User & Between Object 1: the object within the red mask contour, Object 2: the object within the blue mask contour, and Object 3: the object within the green mask contour, which has the most food?
 \\
\midrule
\textbf{Ground Truth } & Object 1 \\ \midrule
GPT-4V & Based on the image, Object 3 (the object within the green mask contour) appears to have the most food. It shows a bowl filled with a larger quantity of broccoli compared to Object 1 (the object within the red mask contour), which has a smaller amount of food, and Object 2 (the object within the blue mask contour), which contains bread, also in a smaller quantity than Object 3.
 \\ \midrule
LLaVA-1.5-13B & Object 2: the object within the blue mask contour, has the most food.
 \\ \midrule
\newshortname{}-7B & Object 1: the object within the red mask contour has the most food.
 \\ \midrule
\newshortname{}-13B & The object within the red mask contour has the most food.
 \\
\bottomrule

\end{tabular}
}
  \end{minipage}
\captionof{table}{Failure cases for GPT-4V on \newbenchmarkshortname{}.}
\label{fig:regionbench-failure4}  
\end{table}

\begin{table}
  \begin{minipage}{0.99\linewidth}
\centering
\scalebox{0.80}{
\begin{tabular}{l p{7.5cm} }

\toprule
\midrule
\textbf{(a)} & \\

&  \includegraphics[height=3.5cm]{figs_regionbench/17.png} \\
User & What is the person marked by the purple point holding?\\
\midrule
\textbf{Ground Truth } & a pair of skis\textless{}OR\textgreater{}skis \\ \midrule
GPT-4V &  The person marked by the purple point appears to be holding a pair of ski poles. Ski poles are used by skiers for balance and propulsion on the snow.
 \\ \midrule
LLaVA-1.5-13B & The person marked by the purple point is holding a green flag.
 \\ \midrule
\newshortname{}-7B & The person marked by the purple point is holding a pair of skis.
 \\ \midrule
\newshortname{}-13B  &  The person marked by the purple point is holding a pair of skis.
 \\
\bottomrule

\textbf{(b)} & \\
&  \includegraphics[height=3.5cm]{figs_regionbench/279.png} \\
User &Between Object 1: the object within the red mask contour, Object 2: the object within the blue mask contour, and Object 3: the object within the green mask contour, which one has something on top of it? \\
\midrule
\textbf{Ground Truth } & Object 2 \\ \midrule
GPT-4V & Object 3, the object within the green mask contour, has something on top of it. It appears to be a coffee pot placed on top of what could be a coffee machine.
  \\ \midrule
LLaVA-1.5-13B & Object 3, which is the coffee maker, has something on top of it.
 \\ \midrule
\newshortname{}-7B & Object 2: the object within the blue mask contour has something on top of it.
 \\ \midrule
\newshortname{}-13B & Object 2: the object within the blue mask contour has something on top of it.
 \\
\bottomrule

\end{tabular}
}
  \end{minipage}
\captionof{table}{Failure cases for GPT-4V on \newbenchmarkshortname{}.}
\label{fig:regionbench-failure5}  
\end{table}


\end{document}